\documentclass{article}
\usepackage{multirow}

\usepackage[nonatbib,preprint]{neurips_2025}

\usepackage[utf8]{inputenc} 
\usepackage[T1]{fontenc}    
\usepackage{hyperref}       
\usepackage{url}            
\usepackage{booktabs}       
\usepackage{amsfonts}       
\usepackage{nicefrac}       
\usepackage{microtype}      
\usepackage{enumitem}
\usepackage[dvipsnames,table,xcdraw]{xcolor}
\usepackage{graphicx}
\usepackage{subfigure}
\usepackage{amsmath, amssymb, amsthm}
\usepackage[numbers]{natbib}

\theoremstyle{plain}
\newtheorem*{theorem*}{Theorem}
\newtheorem{theorem}{Theorem}

\theoremstyle{definition}
\newtheorem{definition}[theorem]{Definition}

\theoremstyle{remark}

\usepackage{algorithm}
\usepackage{algorithmic}

\title{Protein-\texttt{SE(3)}: Benchmarking \texttt{SE(3)}-based Generative Models for Protein Structure Design}

\author{Lang Yu \textsuperscript{1,2, ${\dagger}$}, Zhangyang Gao \textsuperscript{3, ${\dagger}$}, Cheng Tan \textsuperscript{3}, Qin Chen \textsuperscript{1, ${*}$}, Jie Zhou \textsuperscript{1}, Liang He \textsuperscript{1}\\
\textsuperscript{1} School of Computer Science and Technology, East China Normal University\\
\textsuperscript{2} Shanghai Institute of AI for Education, East China Normal University \\
\textsuperscript{3} AI Lab, Research Center for Industries of the Future, Westlake University\\
\thanks{$^{\dagger}$Equal Contribution, $^{*}$Corresponding Author.}
}

\makeatletter
\def\thanks#1{\protected@xdef\@thanks{\@thanks
\protect\footnotetext{#1}}}
\makeatother

\begin{document}

\maketitle

\begin{abstract}
  \texttt{SE(3)}-based generative models have shown great promise in protein geometry modeling and effective structure design. However, the field currently lacks a modularized benchmark to enable comprehensive investigation and fair comparison of different methods. In this paper, we propose Protein-\texttt{SE(3)}, a new benchmark based on a unified training framework, which comprises protein scaffolding tasks, integrated generative models, high-level mathematical abstraction, and diverse evaluation metrics. 
  Recent advanced generative models designed for protein scaffolding, from multiple perspectives like DDPM (Genie1 and Genie2), Score Matching (FrameDiff and RfDiffusion) and Flow Matching (FoldFlow and FrameFlow) are integrated into our framework. All integrated methods are fairly investigated with the same training dataset and evaluation metrics. Furthermore, we provide a high-level abstraction of the mathematical foundations behind the generative models, enabling fast prototyping of future algorithms without reliance on explicit protein structures. 
  Accordingly, we release the first comprehensive benchmark built upon unified training framework for \texttt{SE(3)}-based protein structure design, which is publicly accessible at \href{https://github.com/BruthYU/protein-se3}{\textcolor{blue}{https://github.com/BruthYU/protein-se3}}.
\end{abstract}

\section{Introduction}

The design of protein structures is a fundamental challenge in computational biology, with far-reaching implications like drug discovery and enzyme engineering \cite{cancer_im1,cancer_im2,covid}. Recent advances in AI-driven methods \cite{alphafold2, alphafold3, rfdiffusion, esm} have revolutionized this field, enabling the \textit{de novo} generation of complex, functional proteins. By operating residues in the special Euclidean group $\texttt{SE(3)}=\mathbb{R}^3 \rtimes \texttt{SO(3)}$ and respecting equivariance to rotation and translation, \texttt{SE(3)}-based models demonstrate remarkable quality and diversity in generating protein structures. From multiple perspectives of the diffusion process, researchers have proposed a variety of generative models (DDPM-based \cite{genie1, genie2}, Score Matching-based \cite{framediff, rfdiffusion} and Flow Matching-based \cite{foldflow,frameflow} models) to design protein structures. However, due to differences in their dataset construction and distributed training strategies, it remains challenging to make a fair cross-comparison of these methods. Existing benchmarks like ProteinBench \cite{proteinbench} and Scaffold-Lab \cite{scaffold-lab} primarily focus on the inference performance of aforementioned studies, while overlooking the reproduction and the alignment of the training stages. Furthermore, the implementation of diffusion processes is closely tied to the processing of specific protein data, which hinders the understanding and further development of the underlying mathematical principles.  All these challenges motivate us to establish Protein-\texttt{SE(3)}, a unified benchmark for protein structure design, in which models, metrics and mathematical abstraction are introduced and integrated.

Protein-\texttt{SE(3)} is the first benchmark for protein structure design built upon a unified training framework, aiming to enable fair comparisons across different methods. Backend by Pytorch Lightning \cite{pytorch_lightning}, Protein-\texttt{SE(3)} systematically align different methods with standardized dataset construction and distributed training strategies \cite{pytorch}. Diverse evaluation metric as Quality (scTM, scRMSD), Diversity (Pairwise TM), and Novelty (Max. TM Score to PDB), are also integrated to analyze the strengths and limitations of different methods in protein structure design tasks. 


In addition to the unified training framework and diverse metrics, Protein-\texttt{SE(3)} also abstracts high-level mathematical principles of protein generation models from different perspectives (DDPM, Score Matching and Flow Matching). Based on the synthetic data and wasserstein distance, it enables visualization and analysis of the two marginal diffusion processes in $\mathbb{R}^3$ and \texttt{SO(3)} spaces, facilitating agile prototyping of future algorithms without requiring explicit protein structure data. 

\begin{figure}[ht]
  \centering
  \includegraphics[width=0.9\linewidth]{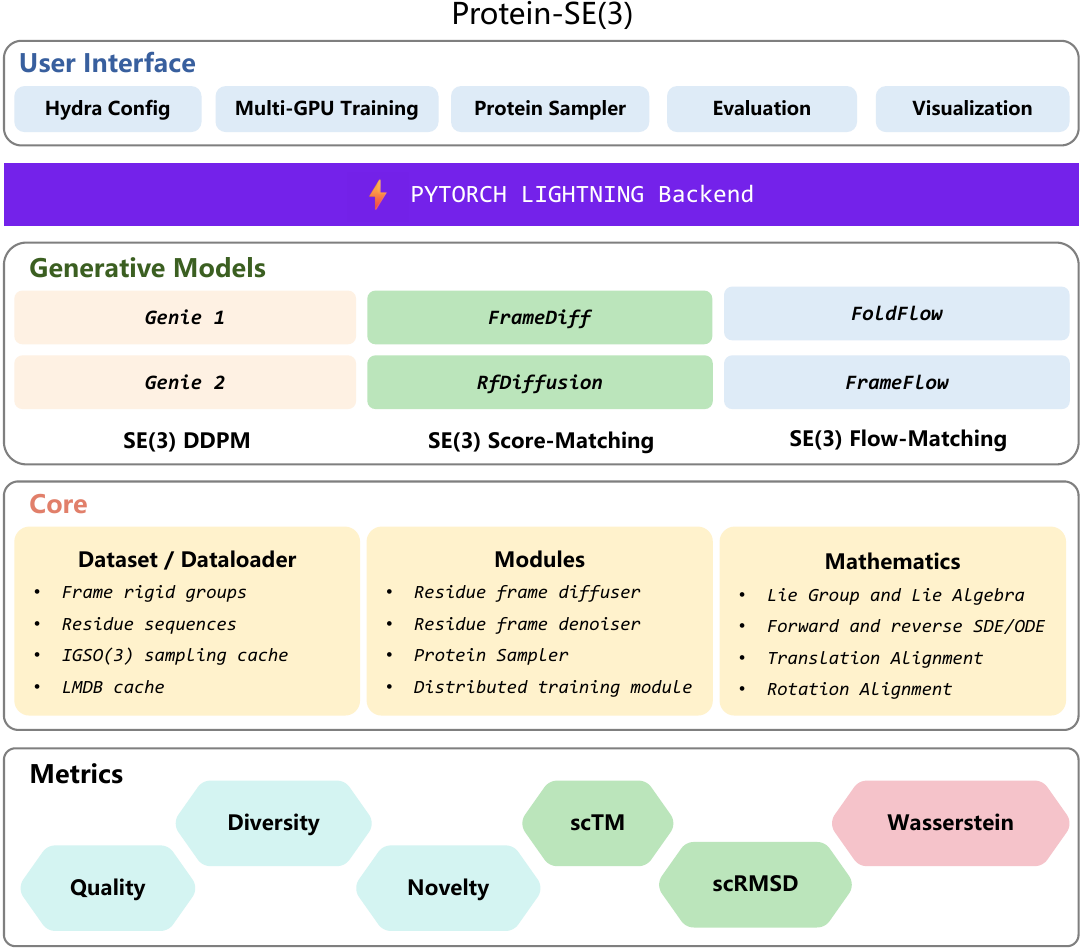}
  \caption{Overview of our proposed Protein-\texttt{SE(3)} benchmark. }
\end{figure}

By supporting a unified training framework, offering high-level mathematical abstractions, and enabling fair cross-comparisons, our proposed benchmark aims to provide researchers with new insights and promote progress in the field. Contributions of Protein-\texttt{SE(3)} are summarized as follows:

\begin{itemize}[leftmargin=1.5em]
\item \textbf{Unified Framework}: We incorporate recently advances in protein structure design into a unified training framework, which enables efficient reproduction and fair cross-comparison

\item \textbf{Mathematical Abstraction}: We provide mathematical abstraction of generative models for protein structure design from different perspectives (DDPM, Score Matching and Flow Matching), facilitating agile prototyping and development on new algorithms without requiring expensive model training on explicit protein structure data.  
\item \textbf{Metrics and Benchmark}: We establish the first benchmark built upon a unified training framework for \texttt{SE(3)}-based protein design, aiming to provide insights  into the strengths and weaknesses of different generative models with a wide range of evaluation metrics.
\end{itemize}

\section{Preliminaries and Problem Formulation}
\begin{figure}[ht]
    \centering

    \subfigure[AlphaFold2 Frame (1AO7.b)]{
    \includegraphics[width=0.23\textwidth]{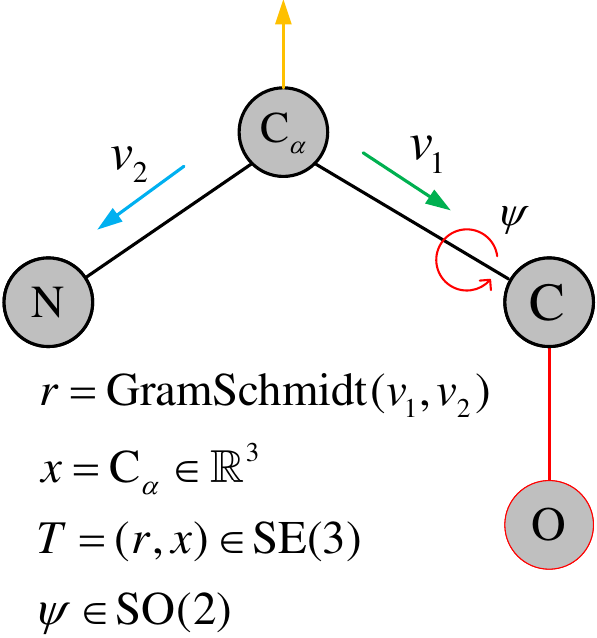}
    \includegraphics[width=0.22\textwidth]{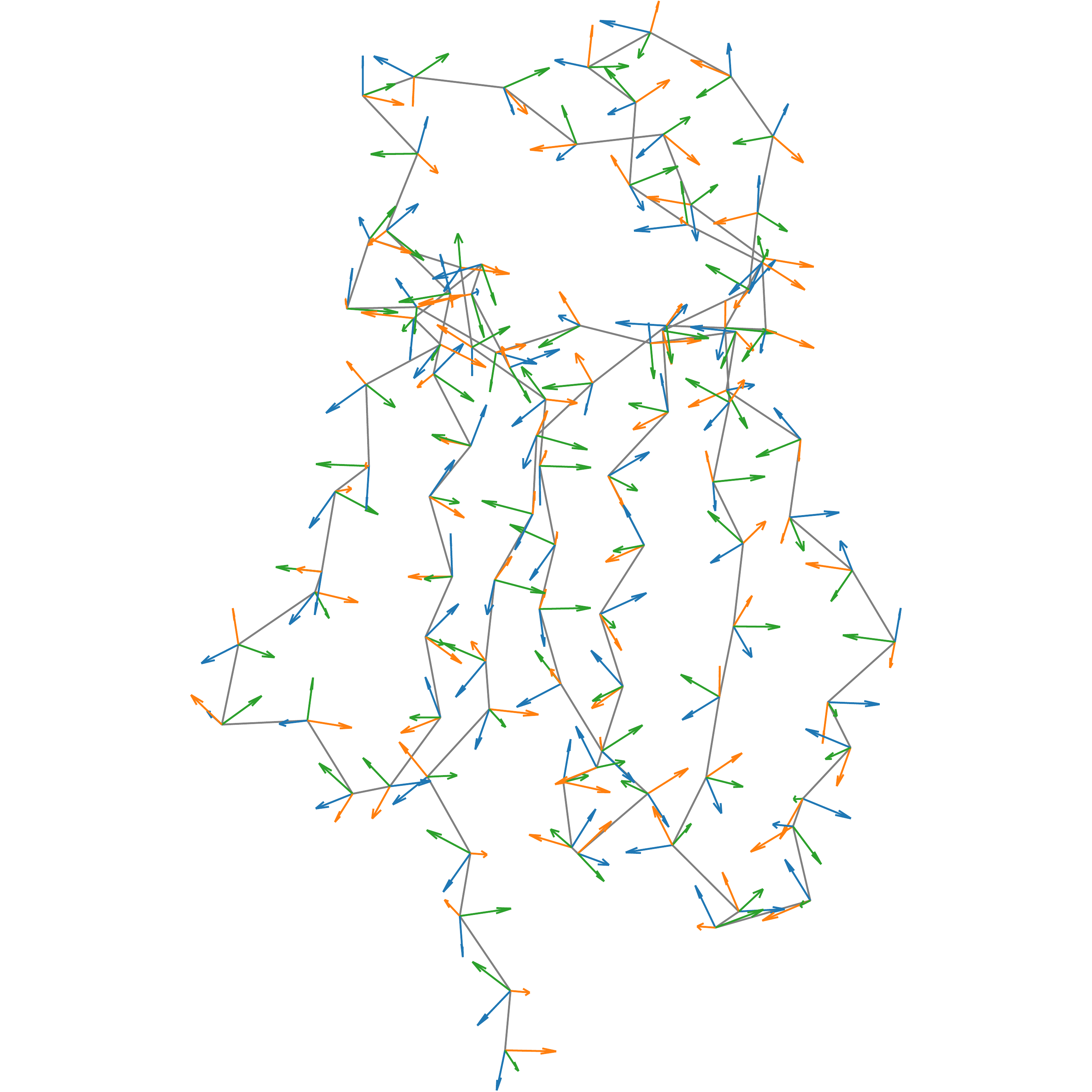}
        \label{formulation1}
    }
    \subfigure[Frenet-Serret Frame (1AO7.b)]{
    \includegraphics[width=0.25\textwidth]{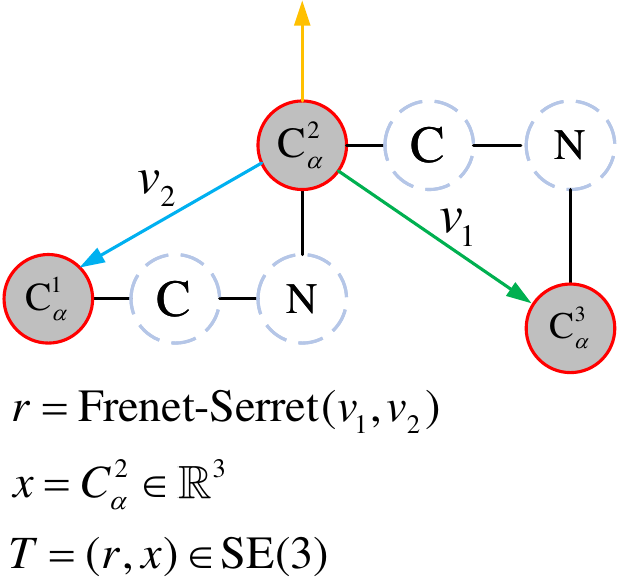}
    \includegraphics[width=0.22\textwidth]{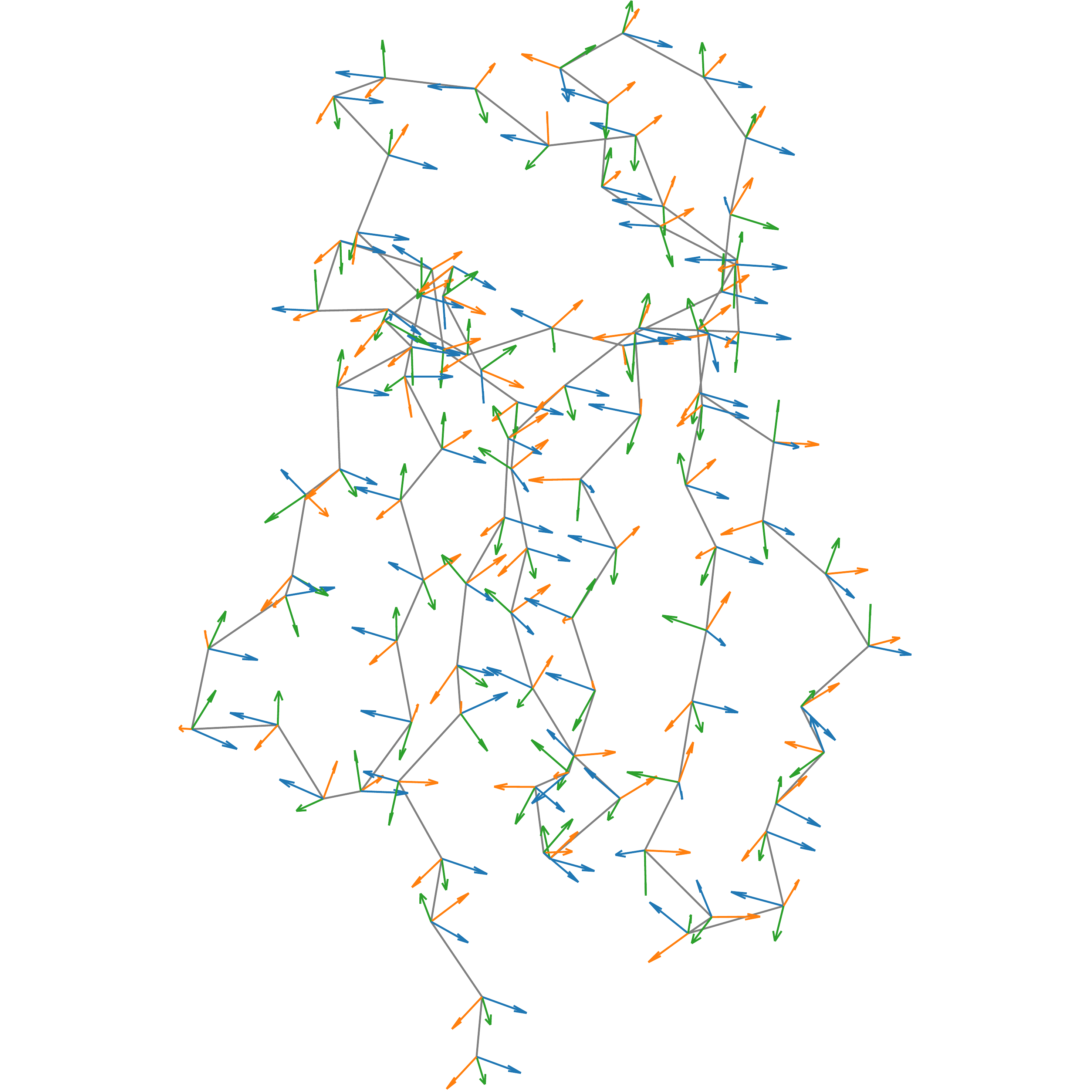}
        \label{formulation2}
    }
    \subfigure[SE(3)-based Protein Structure Generation (Decomposing into \texttt{SO(3)} and $\mathbb{R}^3$)]{
    \includegraphics[width=0.95\textwidth]{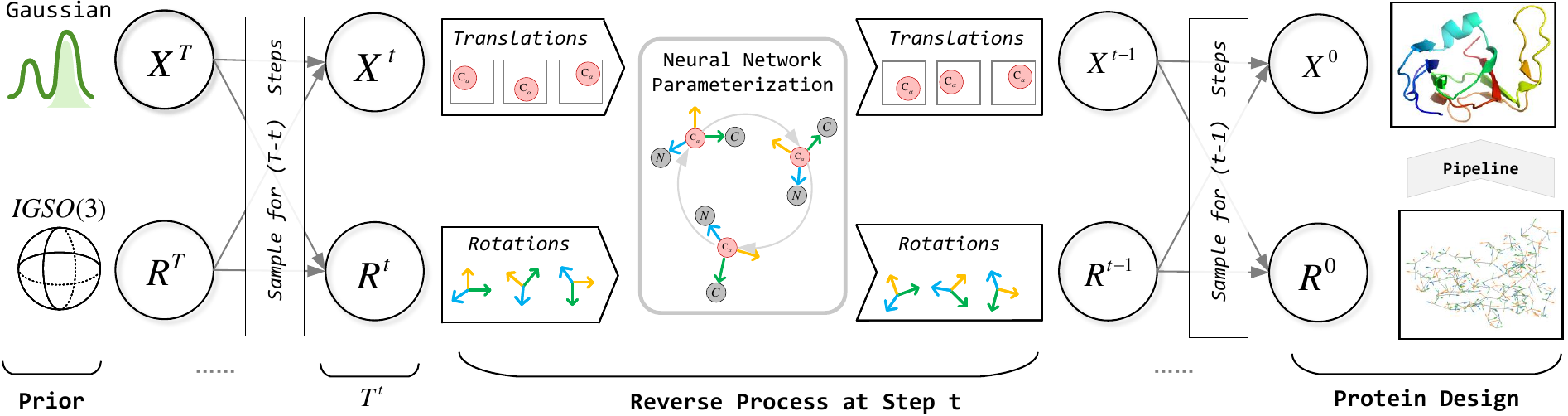}
        \label{formulation3}
    }

    \caption{Protein frame parameterization and problem formulation.}
    \label{formulation}
\end{figure}

\paragraph{Backbone parameterization} As shown in Figures \ref{formulation1} and \ref{formulation2}, the parameterization of the protein backbone mainly follows two paths: the Alphafold2 frame \cite{alphafold2} and the Frenet-Serret frame \cite{frenet1, frenet2}. Generally, each residue is associated with a frame, resulting in $N$ frames that are \texttt{SE(3)}-equivariant for a protein of length $N$: (1) In the
seminal work of AlphaFold2, each frame maps a rigid transformation starting
from idealized coordinates of four heavy atoms $[\text{N}^*, \text{C}_\alpha^*, \text{C}^*, \text{O}^*] \in \mathbb{R}^3$, with $\text{C}_\alpha^*=(0,0,0)$ being centered at the origin. Thus, residue $i\in[1,N]$ is represented as an action $T^i=(r^i,x^i)\in \texttt{SE(3)}$ applied to the idealized frame $[\text{N}, \text{C}_\alpha, \text{C}, \text{O}]^i=T^i\circ [\text{N}^*, \text{C}_\alpha^*, \text{C}^*, \text{O}^*]$. The coordinate of backbone oxygen atom $\text{O}$ is constructed with an additional rotation angel $\varphi$; (2) Another way of backbone parameterization is the Frenet-Serret (FS) frame, which maps each three consecutive $C_\alpha$ into a FS frame. Follwing \cite{genie1,genie2,frenet}, the FS frame $T^i$ is constructed as:
\begin{equation}
t^i=\frac{x^{i+1}-x^i}{||x^{i+1}-x^i||},b^i=\frac{t^{i-1}\times t^i}{||t^{i-1}\times t^i||},n^i=b^i\times t^i; \quad r^i=[t^i,b^i,n^i], T^i=(r^i,x^i)
\end{equation}
where the coordinate of the second element is recognized as the translation vector $x^i$. 

\paragraph{Decomposing \texttt{SE(3)} into \texttt{SO(3)} and $\mathbb{R}^3$} To construct the diffusion process (or flow) of \texttt{SE(3)}, the definitions of an inner product and a metric on \texttt{SE(3)} are required to obtain a Riemannian structure \cite{foldflow,framediff}. The common choices in previous studies \cite{framediff, foldflow} are:
\begin{equation}
    \begin{aligned}
    &\text{Inner Products:} \quad \langle r,r^\prime\rangle_{\texttt{SO(3)}}=tr(rr^{\prime T})/2\quad \text{and} \quad \langle x,x^\prime\rangle=\sum\nolimits_{i=1}^3x_ix^\prime_i\\
    &\text{Metric on \texttt{SE(3)}:}\quad \langle(r,x),(r^\prime,x^\prime) \rangle_{\texttt{SE(3)}}=\langle r,r^\prime\rangle_{\texttt{SO(3)}}+\langle x,x^\prime\rangle_{\mathbb{R}^3}
    \end{aligned}
\end{equation}
As shown in Figure \ref{formulation3}, the diffusion (or flow) of \texttt{SE(3)} can be decomposed into \texttt{SO(3)} and $\mathbb{R}^3$. 
\begin{definition} \texttt{IGSO(3)}: The Isotropic Gaussian Distribution on \texttt{SO(3)} is a probability distribution over the 3D rotation group \texttt{SO(3)}, which generalizes the idea of a Gaussian (normal) distribution to the non-Euclidean manifold of 3D rotations \cite{igso3}. 

\end{definition}

The structure design task is to learn the reverse process from the empirical  distributions (Gaussian for $\mathbb{R}^3$ and \texttt{IGSO(3)} for \texttt{SO(3)}) to the target structure distributions of actual proteins. 


\section{Mathematical Abstraction}
\label{mathematical}

\texttt{SE(3)}-based methods model and align protein structures on both the translational $\mathbb{R}^3$ and rotational \texttt{SO(3)} spaces. To abstract and formalize the mathematical principles underlying protein structure design models, we developed a dedicated mathematical toolkit to visualize and analyze the distribution alignment process with the measurement of 1st-order Wasserstein distance. This toolkit is built upon simple MLP layers training on the synthetic data (3D coordinates and rotation matrices represented as Euler angles), enabling systematic study on how various generative models (DDPM~\cite{ddpm, so3_ddpm}, Score Matching~\cite{score-matching}, and Flow Matching~\cite{foldflow}) align distributions in both the translational and rotational spaces. With our developed toolkit, researchers could efficiently prototype recent progress of generative models without expensive model training on protein dataset.

\begin{figure}[ht]
    \centering

    \subfigure[$\mathbb{R}^3$ alignment on target distribution A]{
    \includegraphics[width=0.19\textwidth]{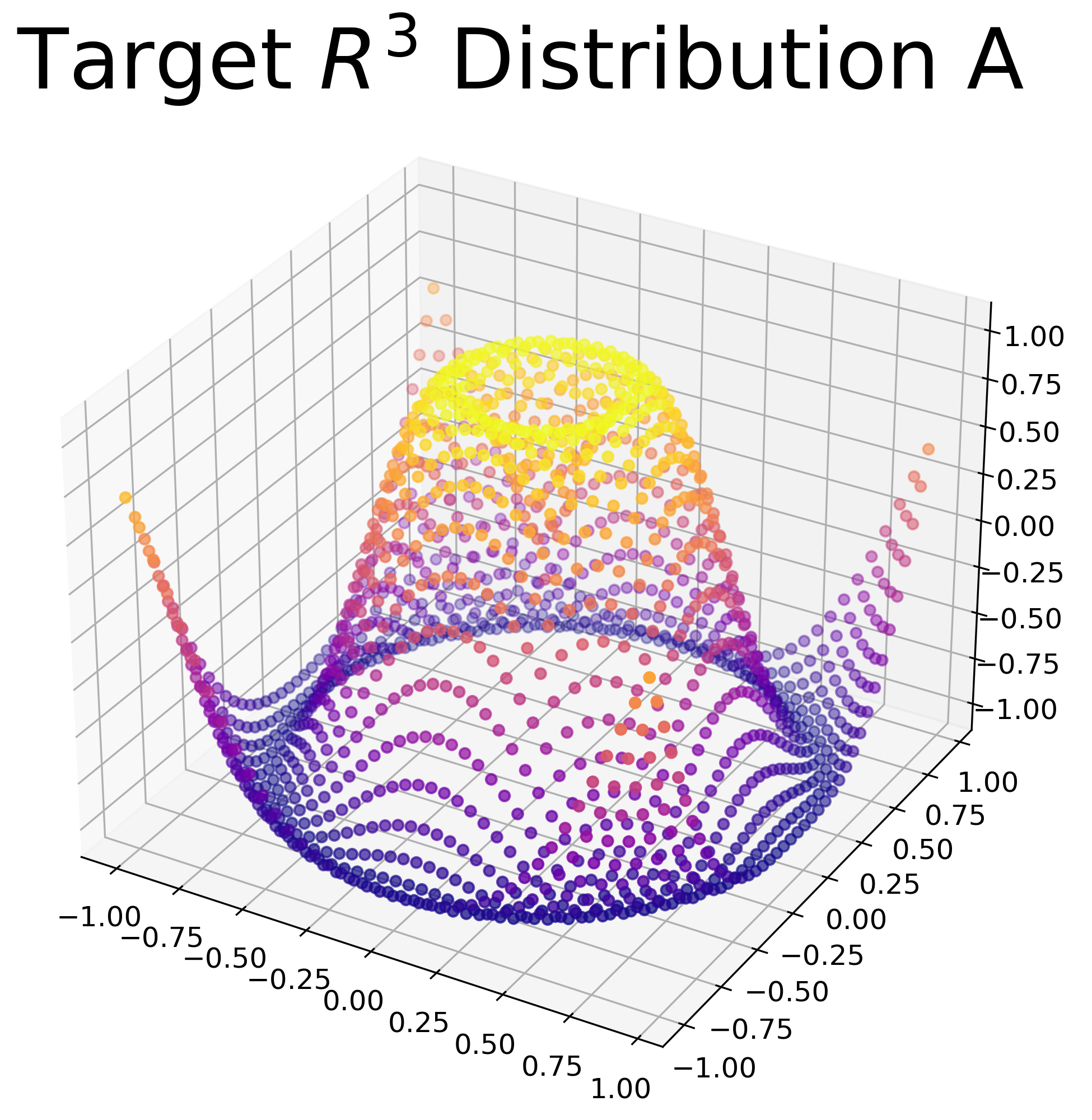}
    \includegraphics[width=0.19\textwidth]{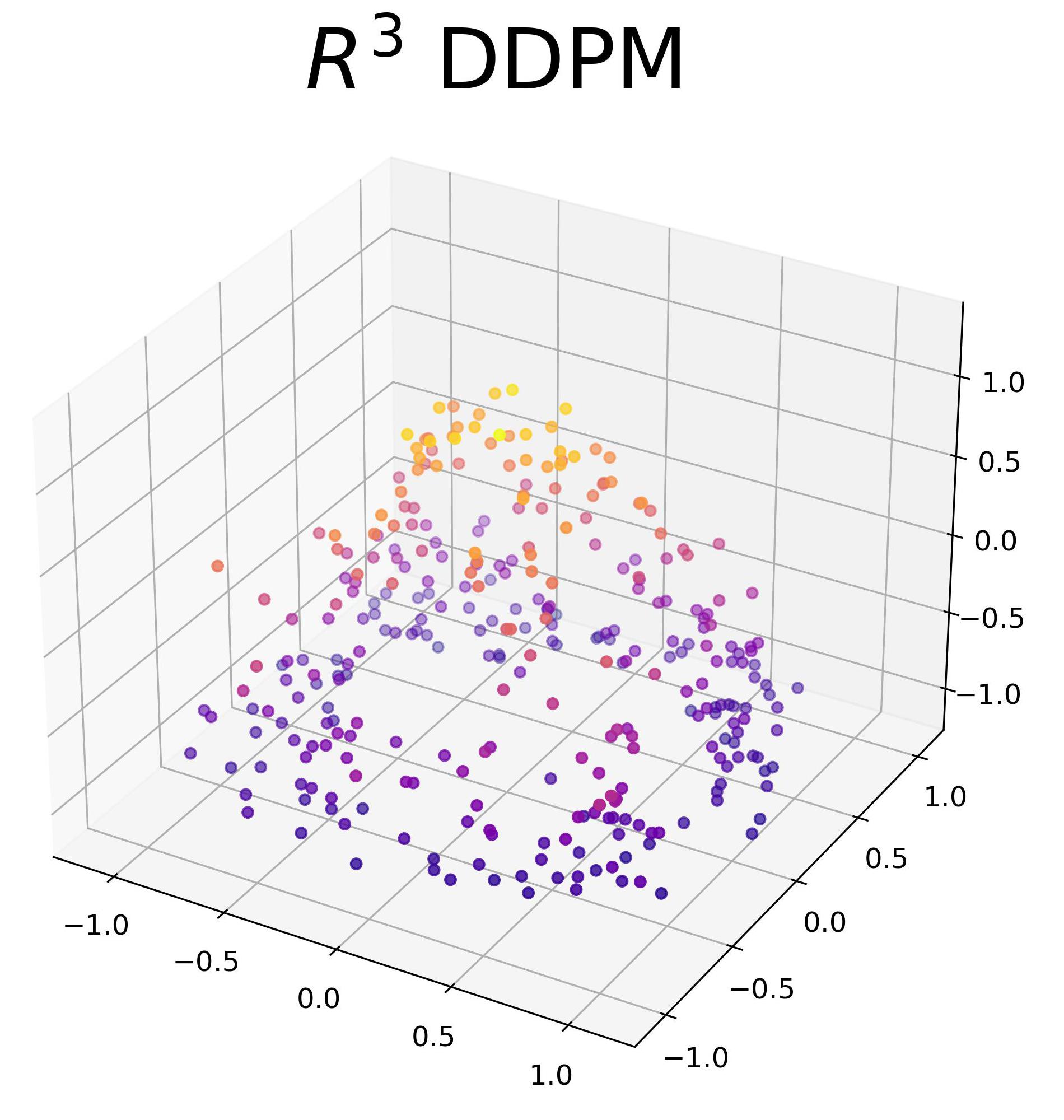}
    \includegraphics[width=0.19\textwidth]{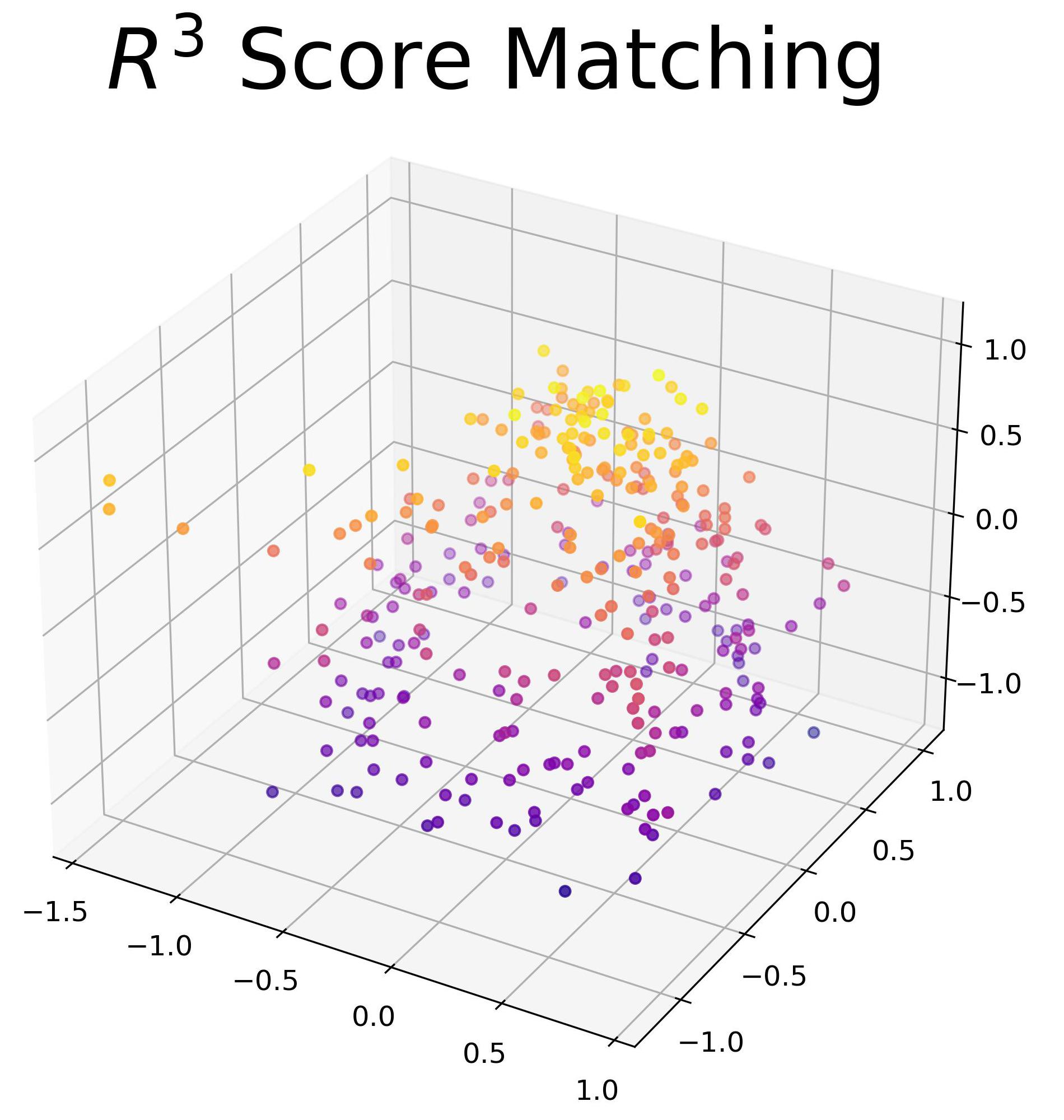}
    \includegraphics[width=0.19\textwidth]{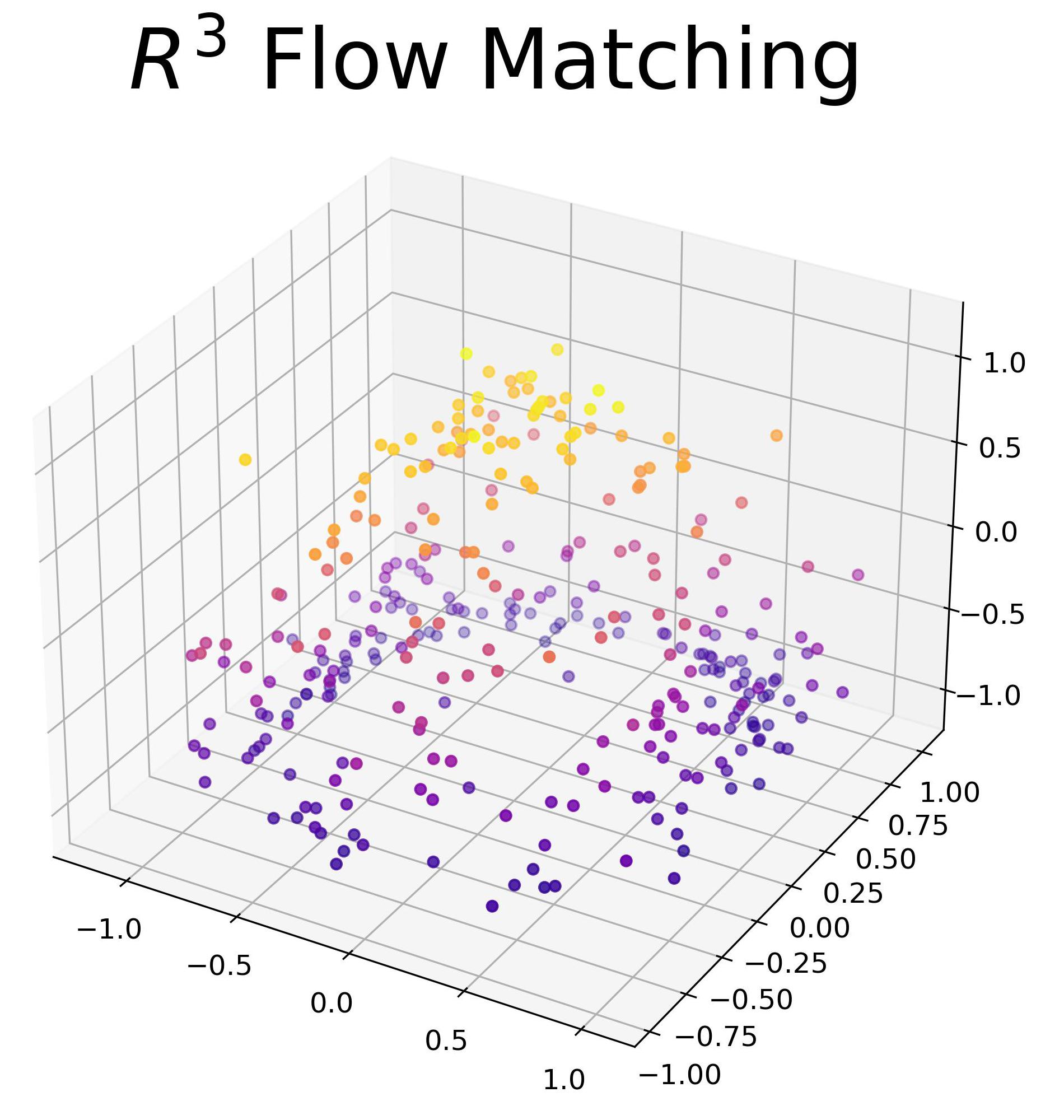}}

    \subfigure[$\mathbb{R}^3$ alignment on target distribution B]{
    \includegraphics[width=0.19\textwidth]{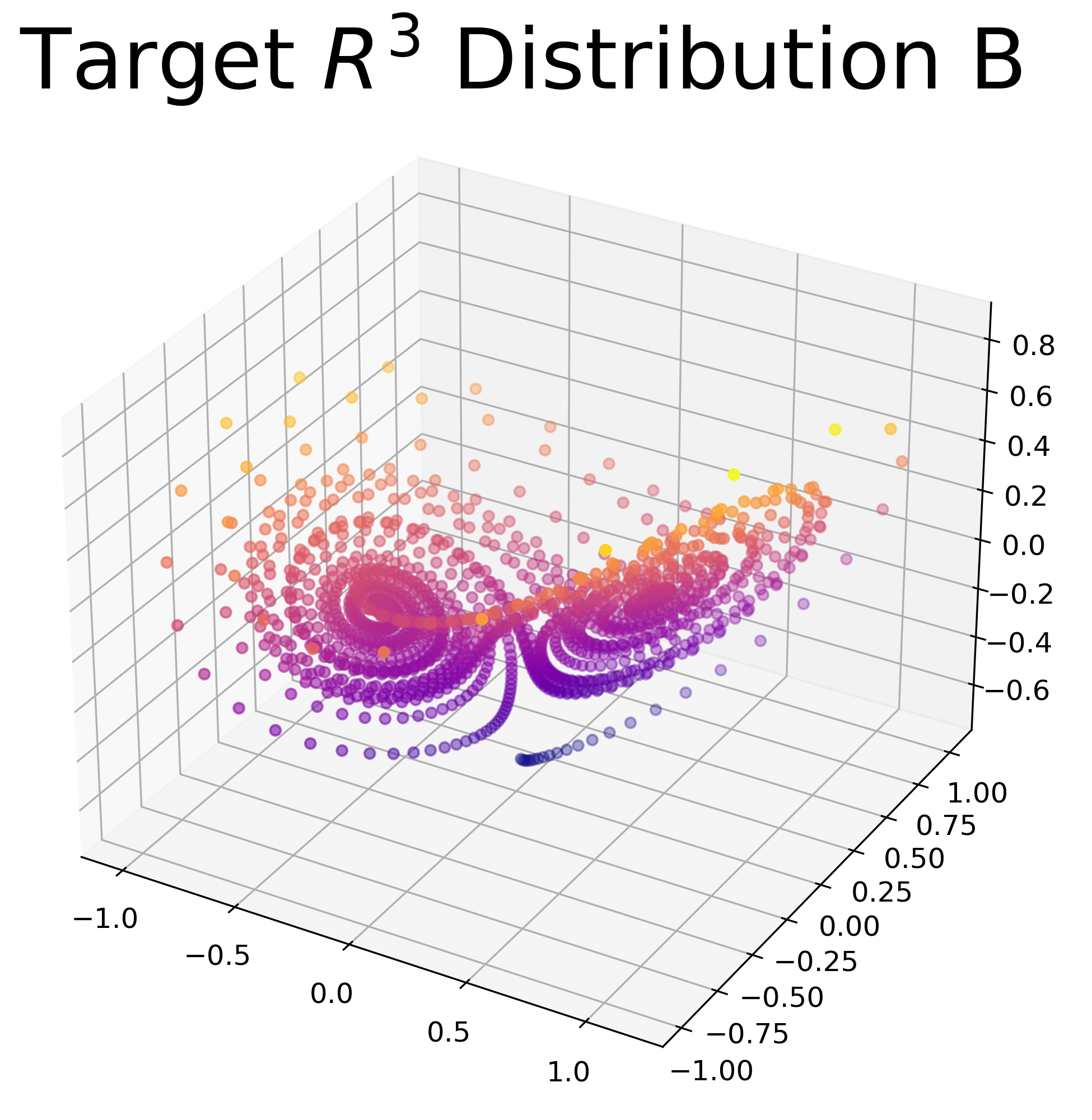}
    \includegraphics[width=0.19\textwidth]{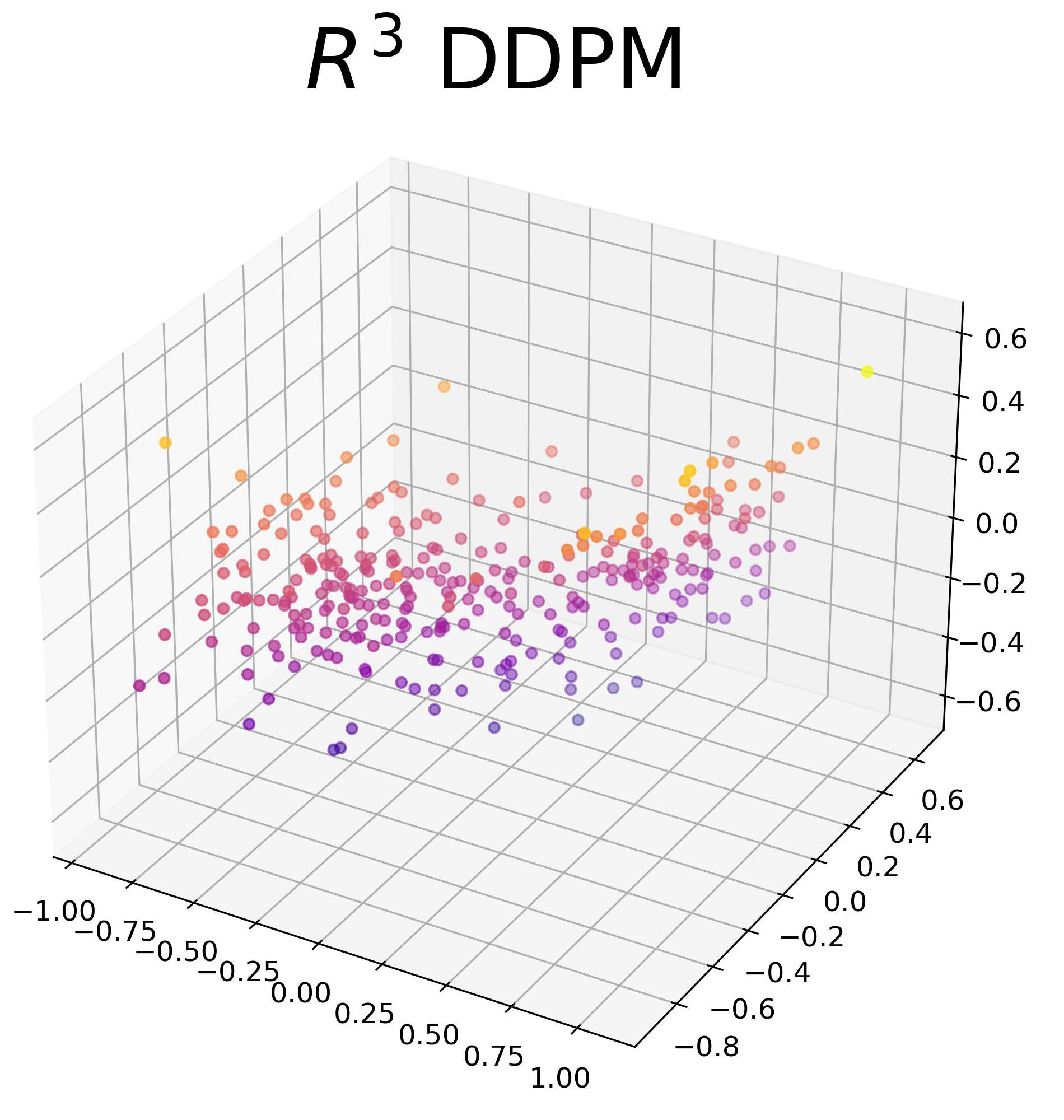}
    \includegraphics[width=0.19\textwidth]{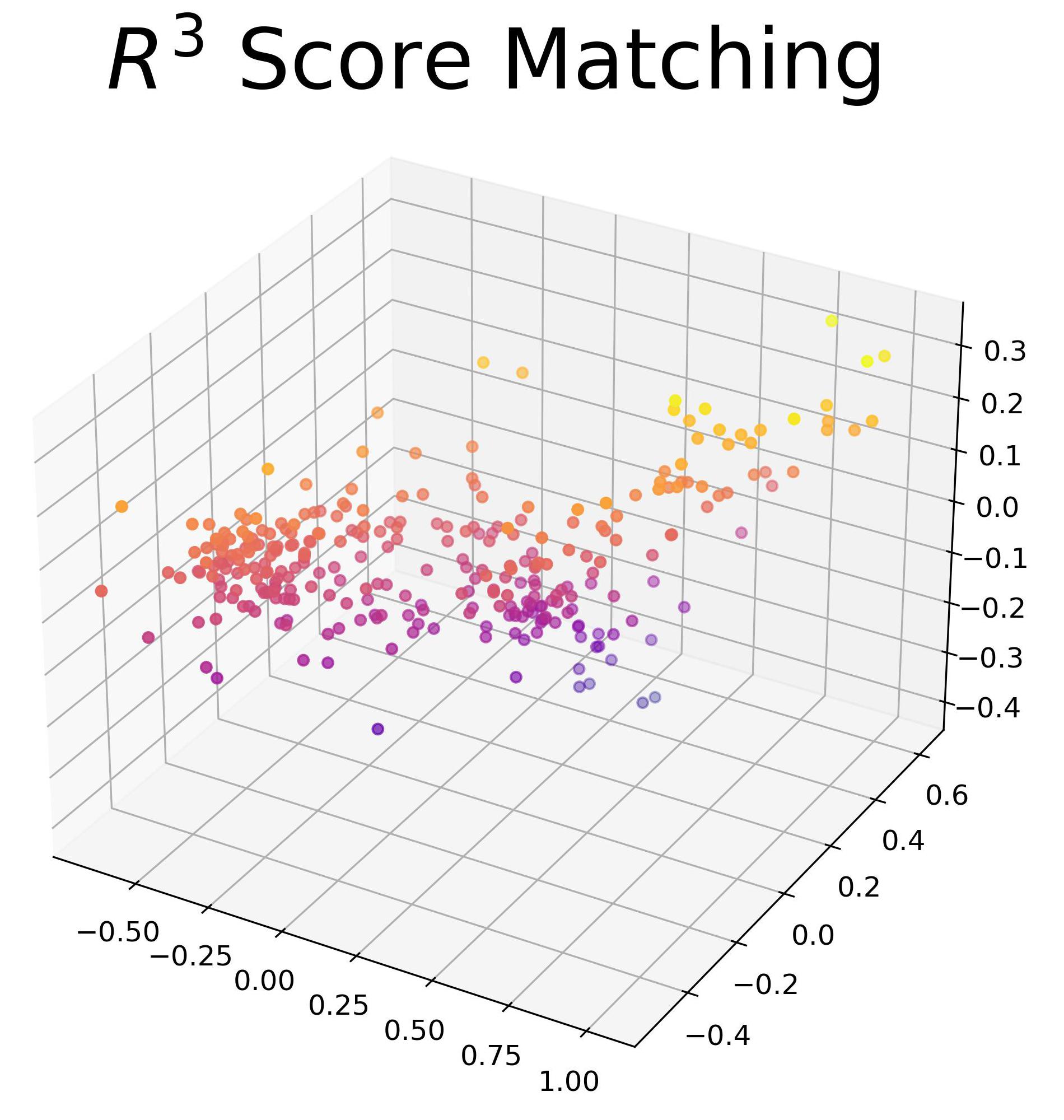}
    \includegraphics[width=0.19\textwidth]{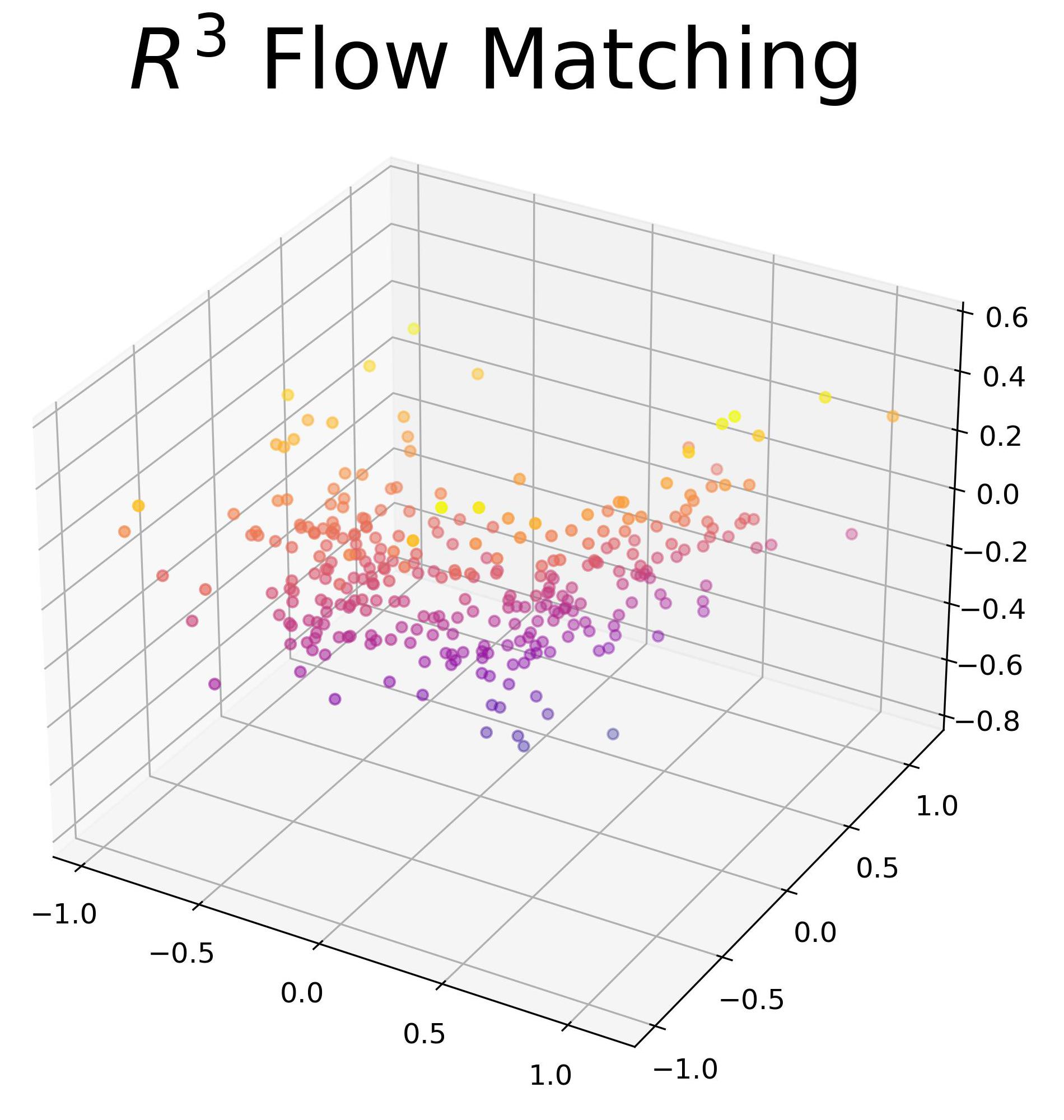}}
    
    \subfigure[Process of $\mathbb{R}^3$ alignment on target A]{
    \includegraphics[width=0.38\textwidth]{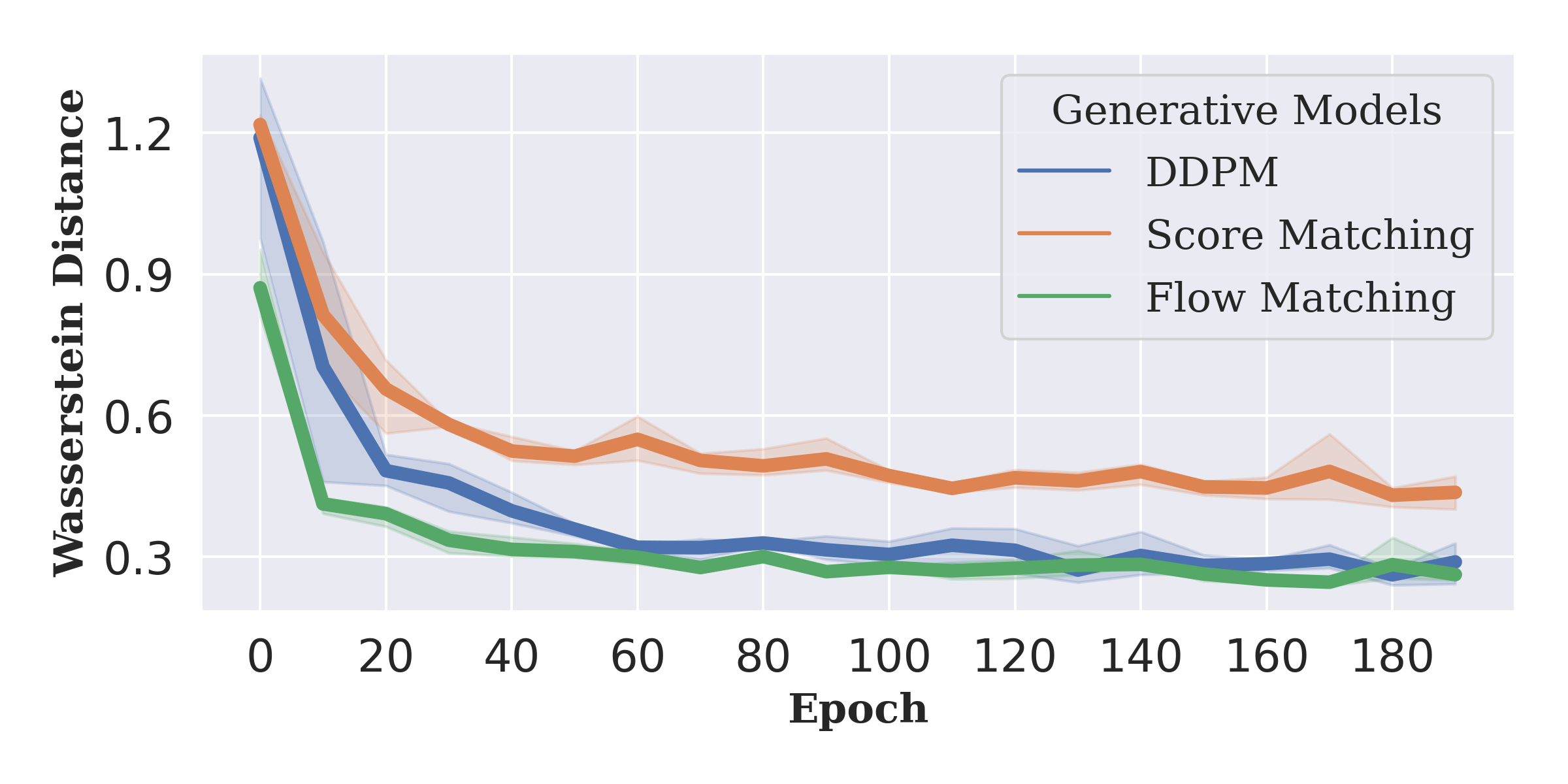}
    }
    \subfigure[Process of $\mathbb{R}^3$ alignment on target B]{
    \includegraphics[width=0.38\textwidth]{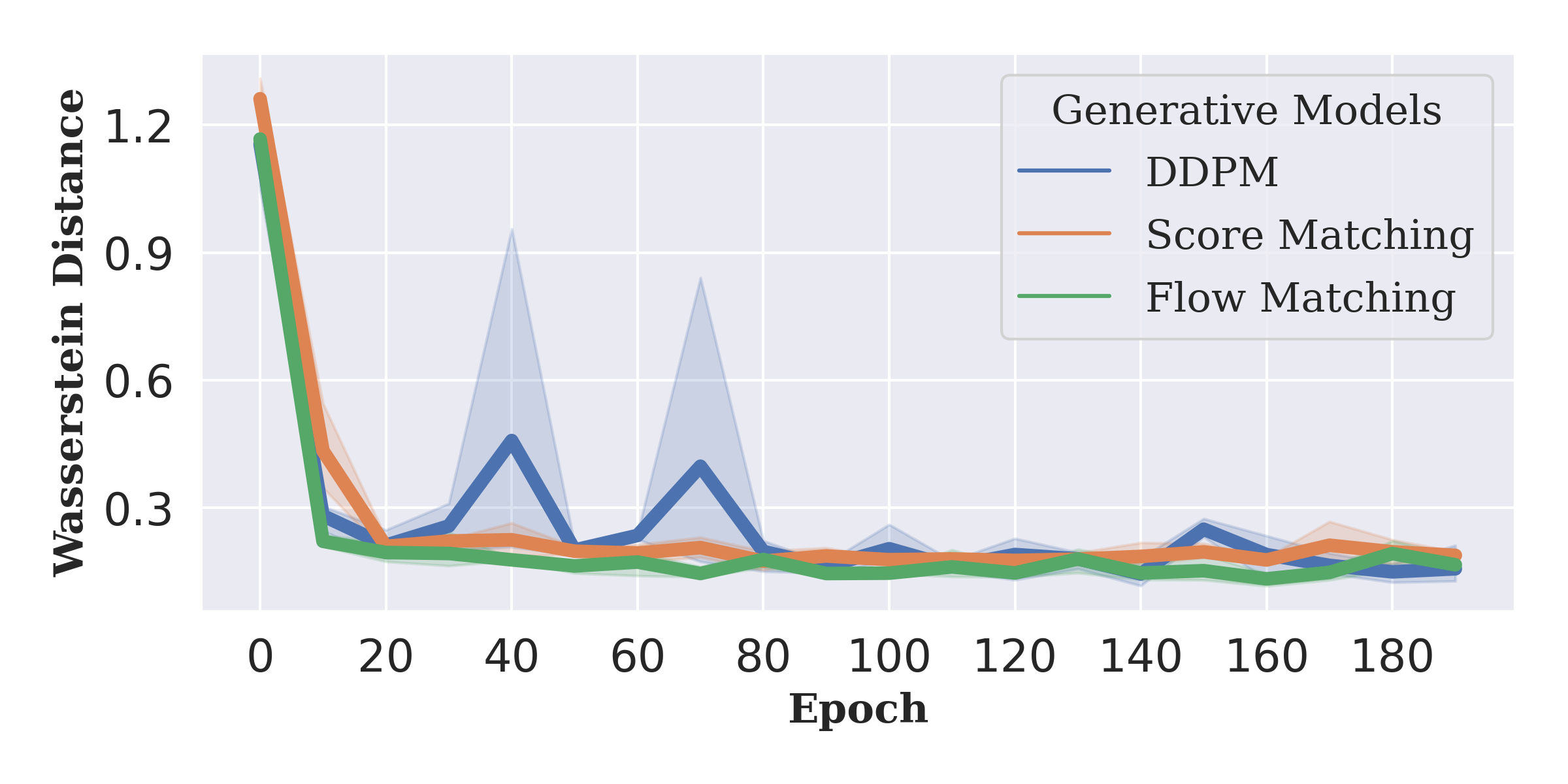}
    }

    \caption{Experiments on $\mathbb{R}^3$ Alignment with different generative modeling methods.}
    \label{translation}
\end{figure}

\begin{definition}Wasserstein Distance: The Wasserstein distance \cite{wasserstein1, wasserstein2} measures the minimum cost required to transform one probability distribution into another. 
For probability distributions $\mu$ and $\upsilon$ over a metric space $\chi$ (e.g. $\mathbb{R}^3$ and \texttt{SO(3)}), let $\Gamma(\mu,\upsilon)$ is the set of all joint distributions on $\chi \times \chi$, the 1st-order Wasserstein distance is defined as:

\begin{equation}
W_1(\mu, \upsilon) = \inf_{\gamma \in \Gamma(\mu, \upsilon)} \mathbb{E}_{(x,y) \sim \gamma} [\|x - y\|]
\label{eq:1}
\end{equation}
where $\gamma(x, y)$ represents a transport plan and $\|x - y\|$ denotes the transport cost from point $x$ to $y$. 

\end{definition}

\paragraph{Translation alignment in $\mathbb{R}^3$ space}
The translations ($C_\alpha$ positions)  of proteins are defined in the standard $\mathbb{R}^3$ 
 Euclidean space, where the diffusion process (or flow) can be accomplished easily through the previously derived closed form equations~\cite{point_cloud}. 
Detailed formulations of translation alignment, as derived from different generative models, are presented in \textcolor{blue}{Appendix. A}. While figure \ref{translation} provides a visualization of the translation alignment: as training epochs increase, the $\mathbb{R}^3$ distributions sampled by three generative models become progressively closer to the target distribution, as evidenced by the decreasing Wasserstein Distance, ultimately achieving alignment with the target distribution.

\begin{figure}[ht]
    \centering

    \subfigure[\texttt{SO(3)} alignment on target distribution A (Euler Angle Representation)]{
    \includegraphics[width=0.19\textwidth]{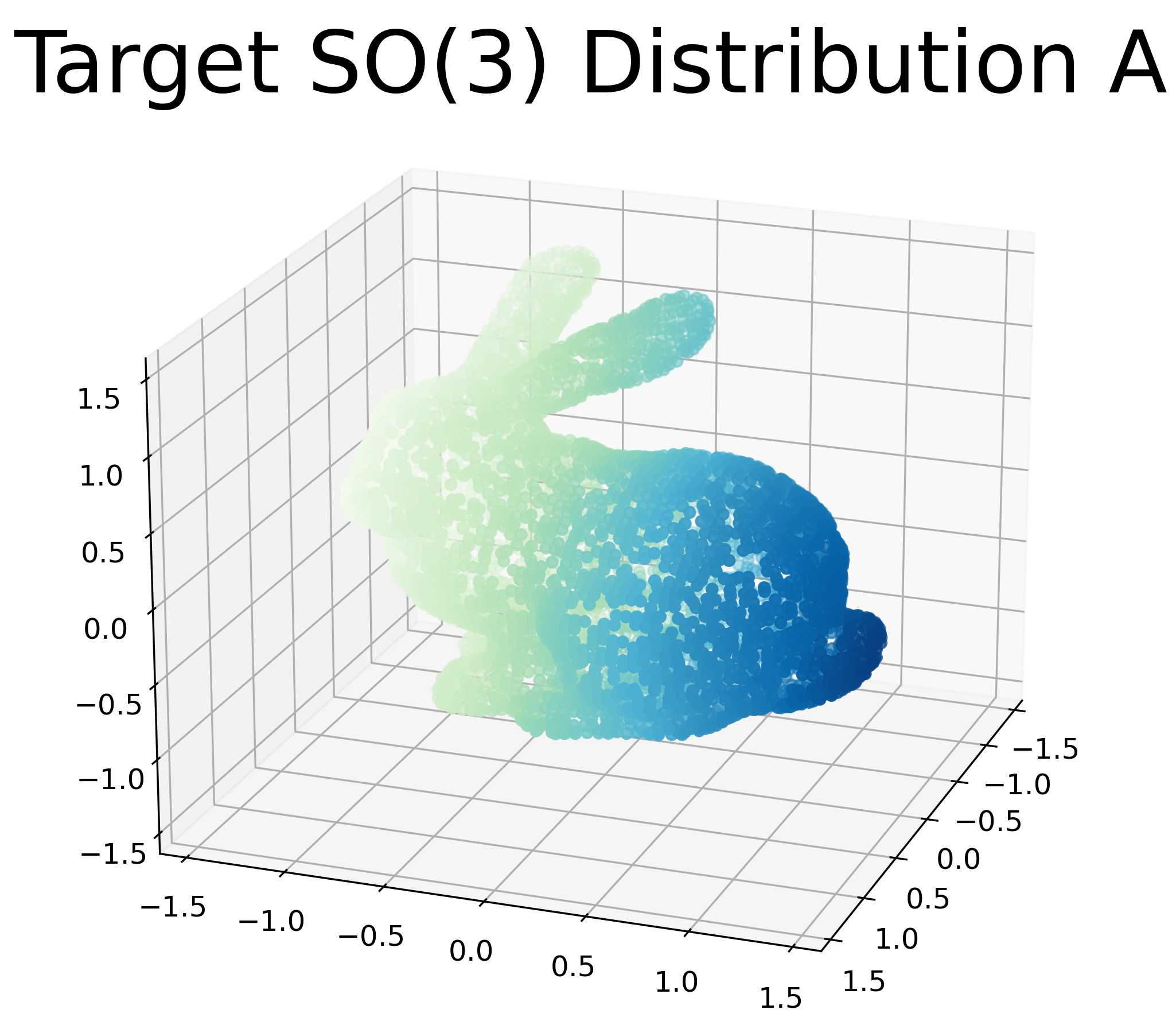}
    \includegraphics[width=0.19\textwidth]{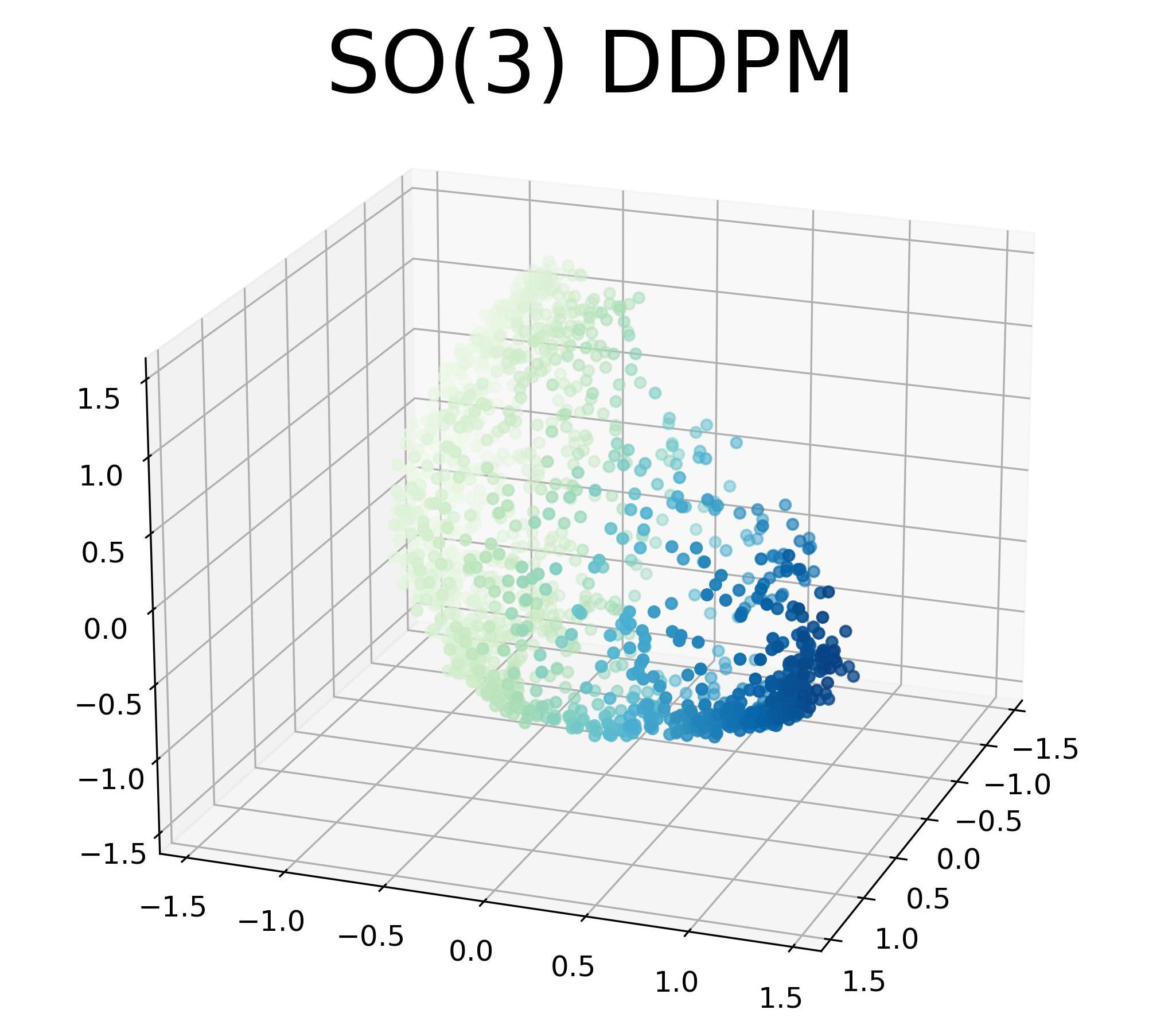}
    \includegraphics[width=0.19\textwidth]{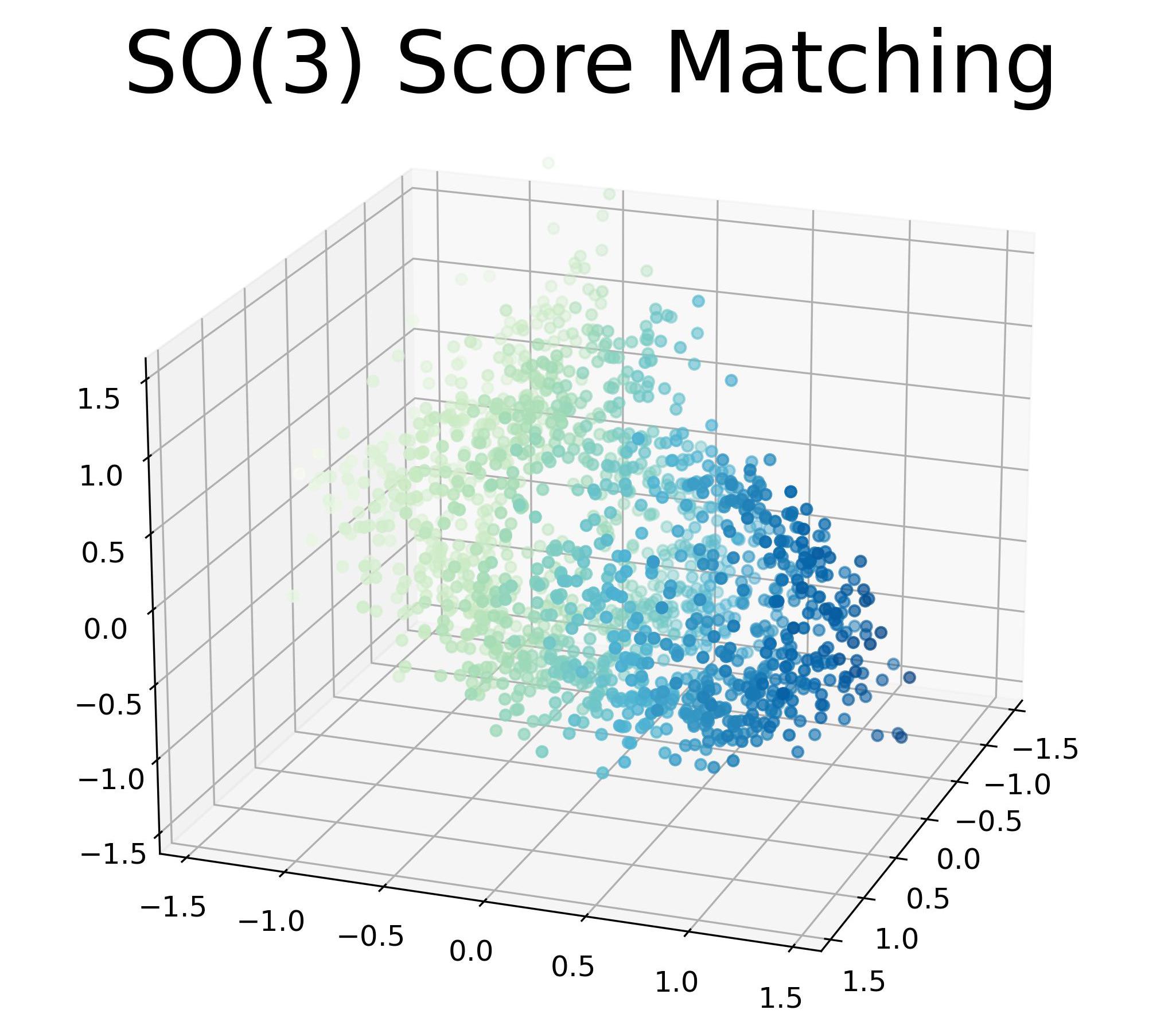}
    \includegraphics[width=0.19\textwidth]{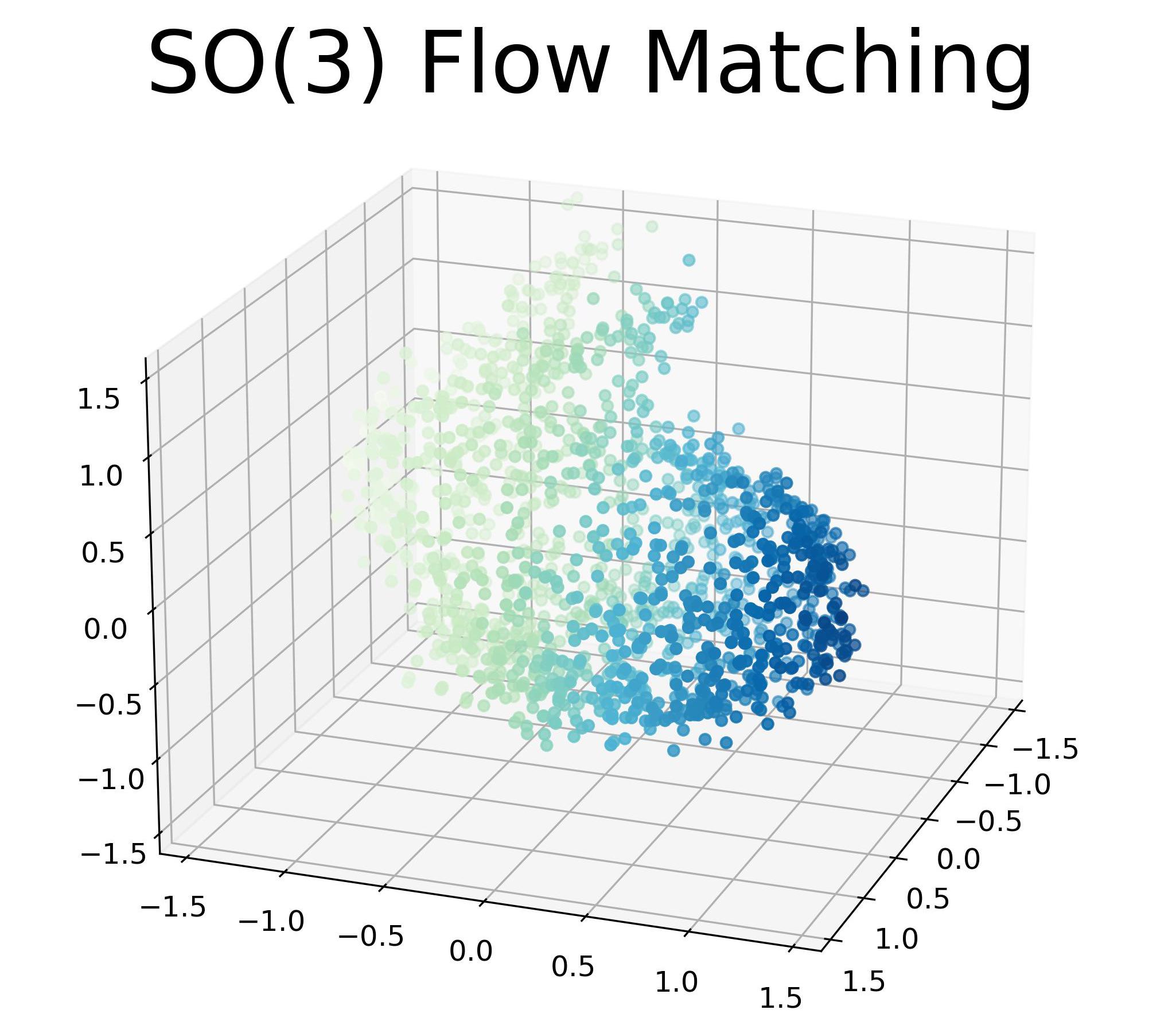}}

    \subfigure[\texttt{SO(3)} alignment on target distribution B (Euler Angle Representation)]{
    \includegraphics[width=0.19\textwidth]{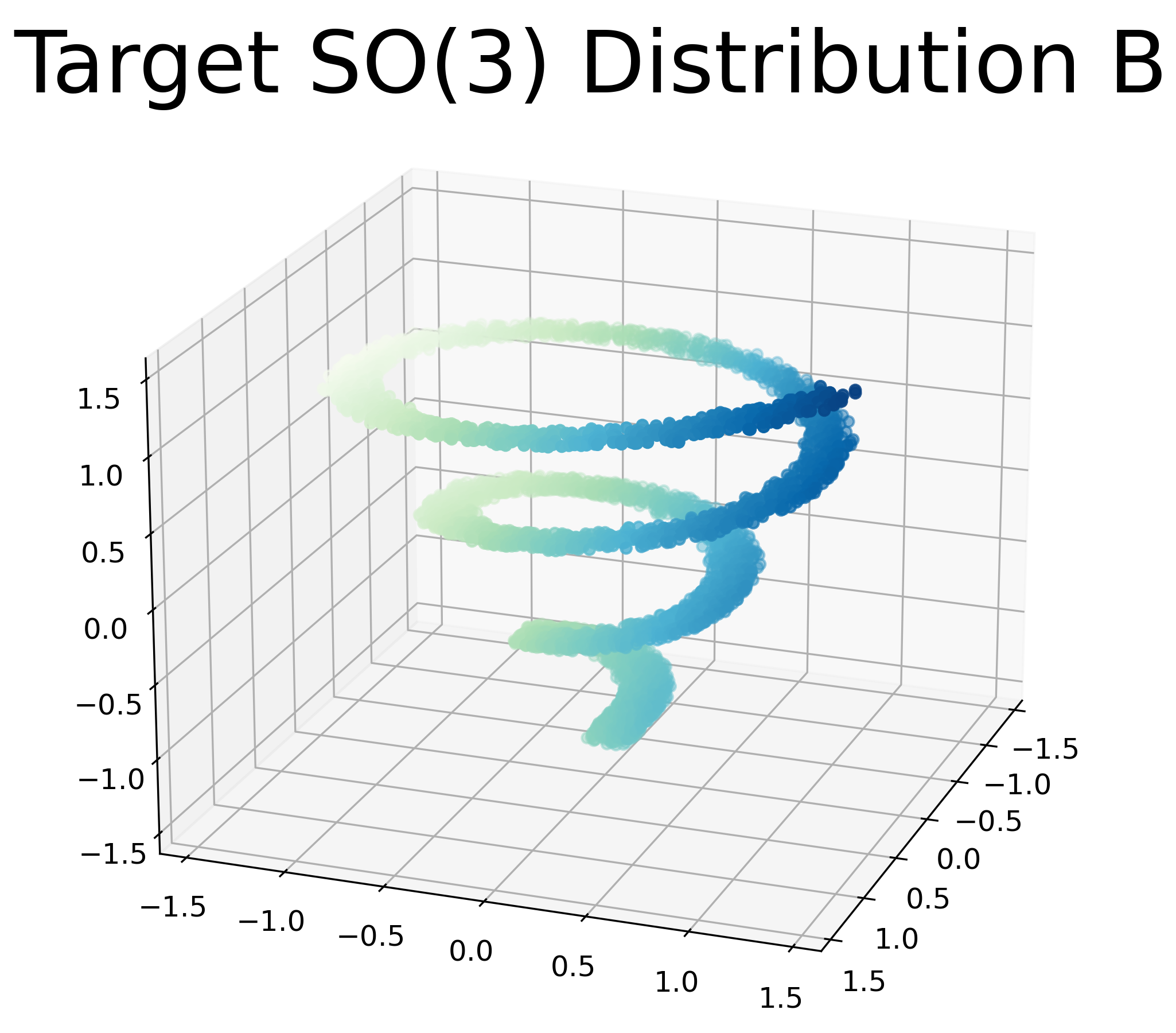}
    \includegraphics[width=0.19\textwidth]{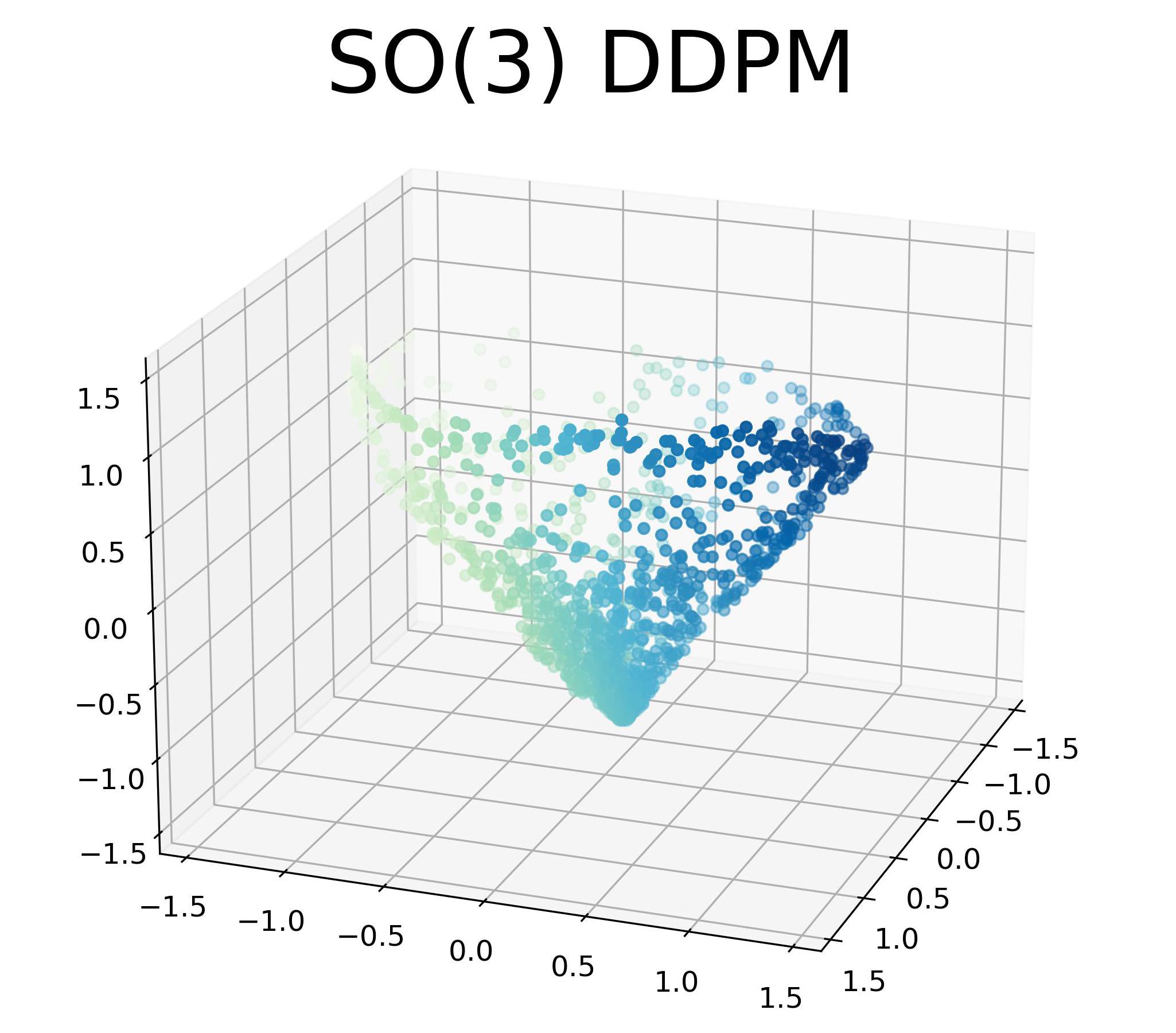}
    \includegraphics[width=0.19\textwidth]{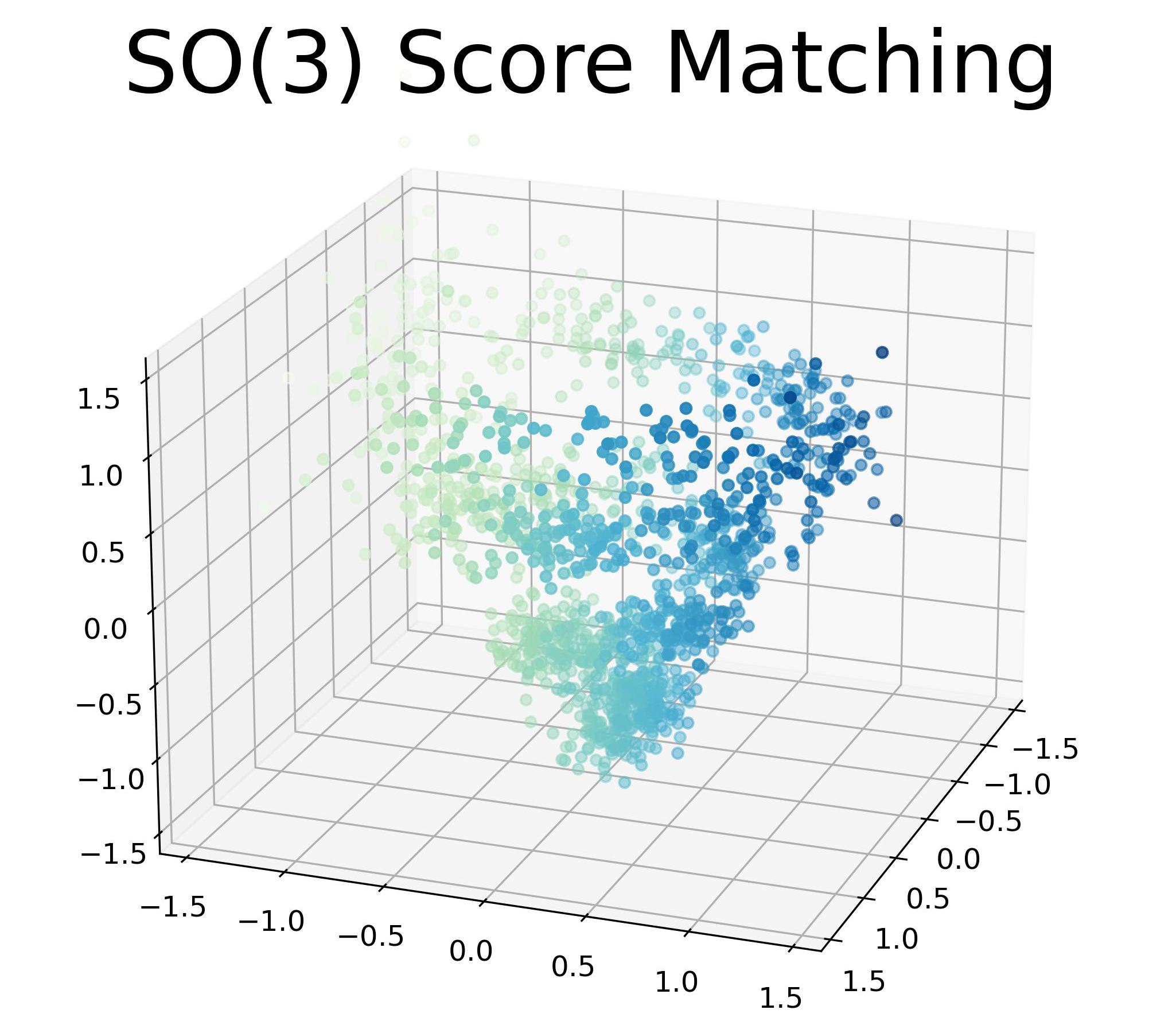}
    \includegraphics[width=0.19\textwidth]{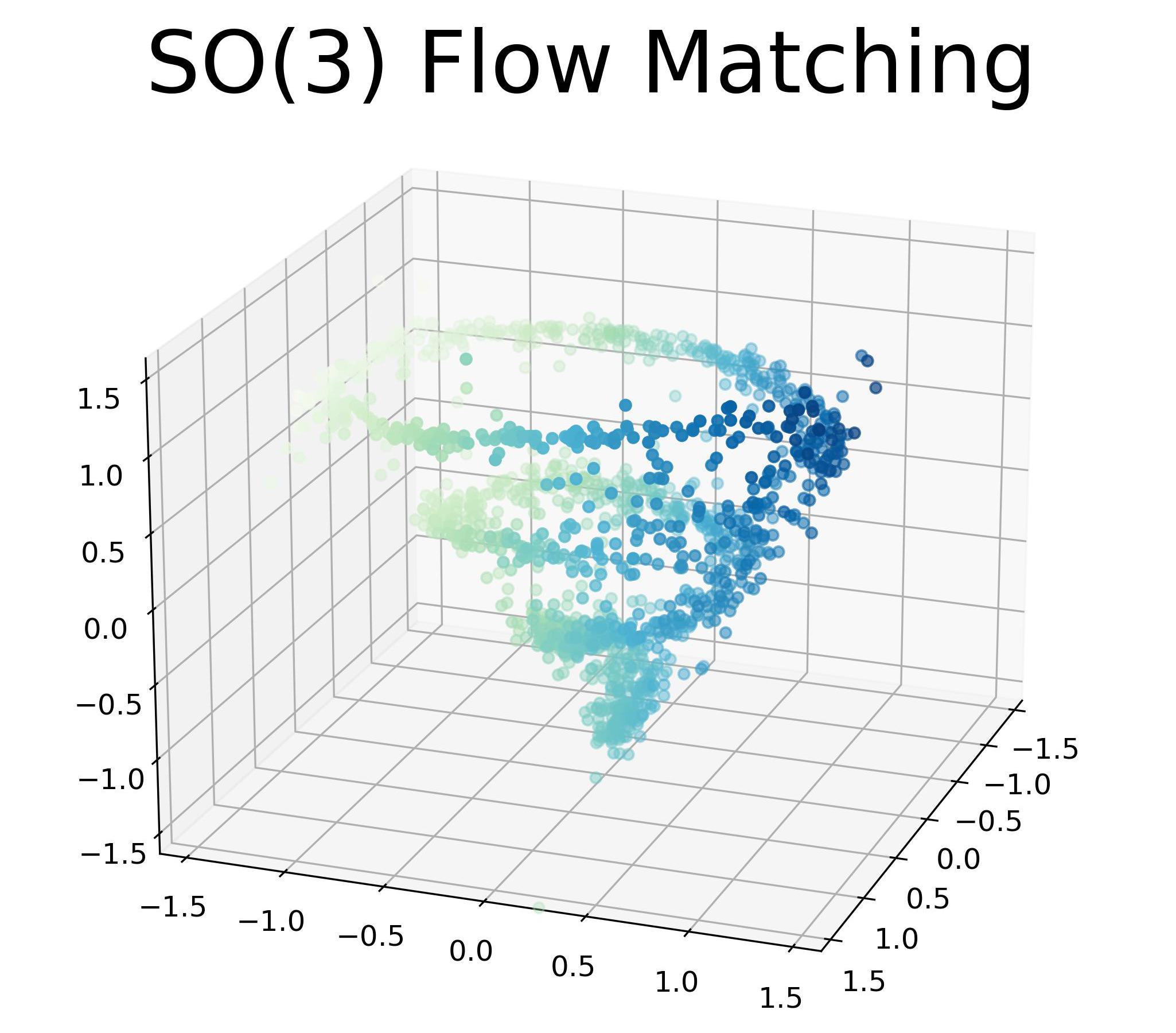}}
    
    \subfigure[Process of \texttt{SO(3)} alignment on target A]{
    \includegraphics[width=0.38\textwidth]{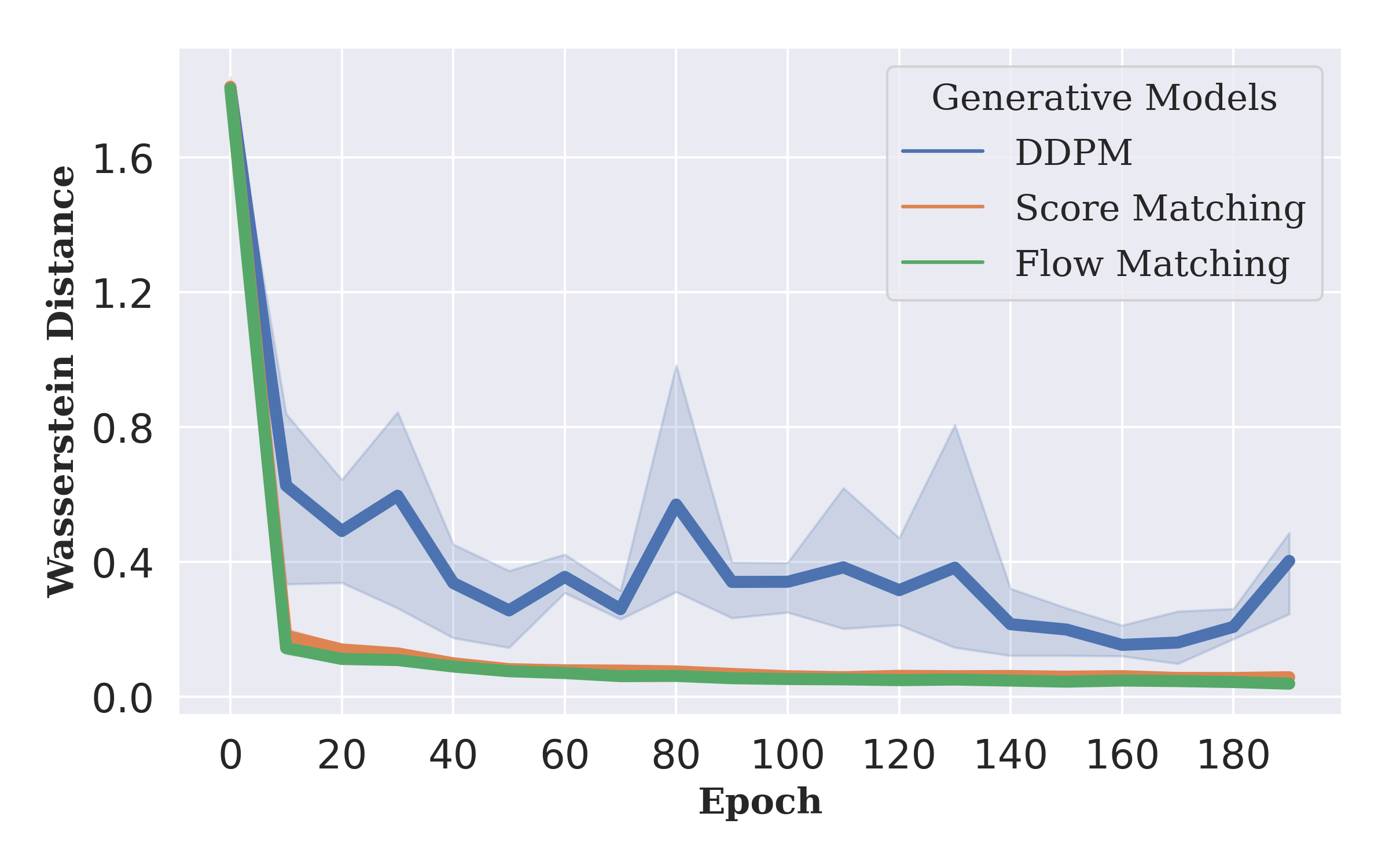}
    }
    \subfigure[Process of \texttt{SO(3)} alignment on target B]{
    \includegraphics[width=0.38\textwidth]{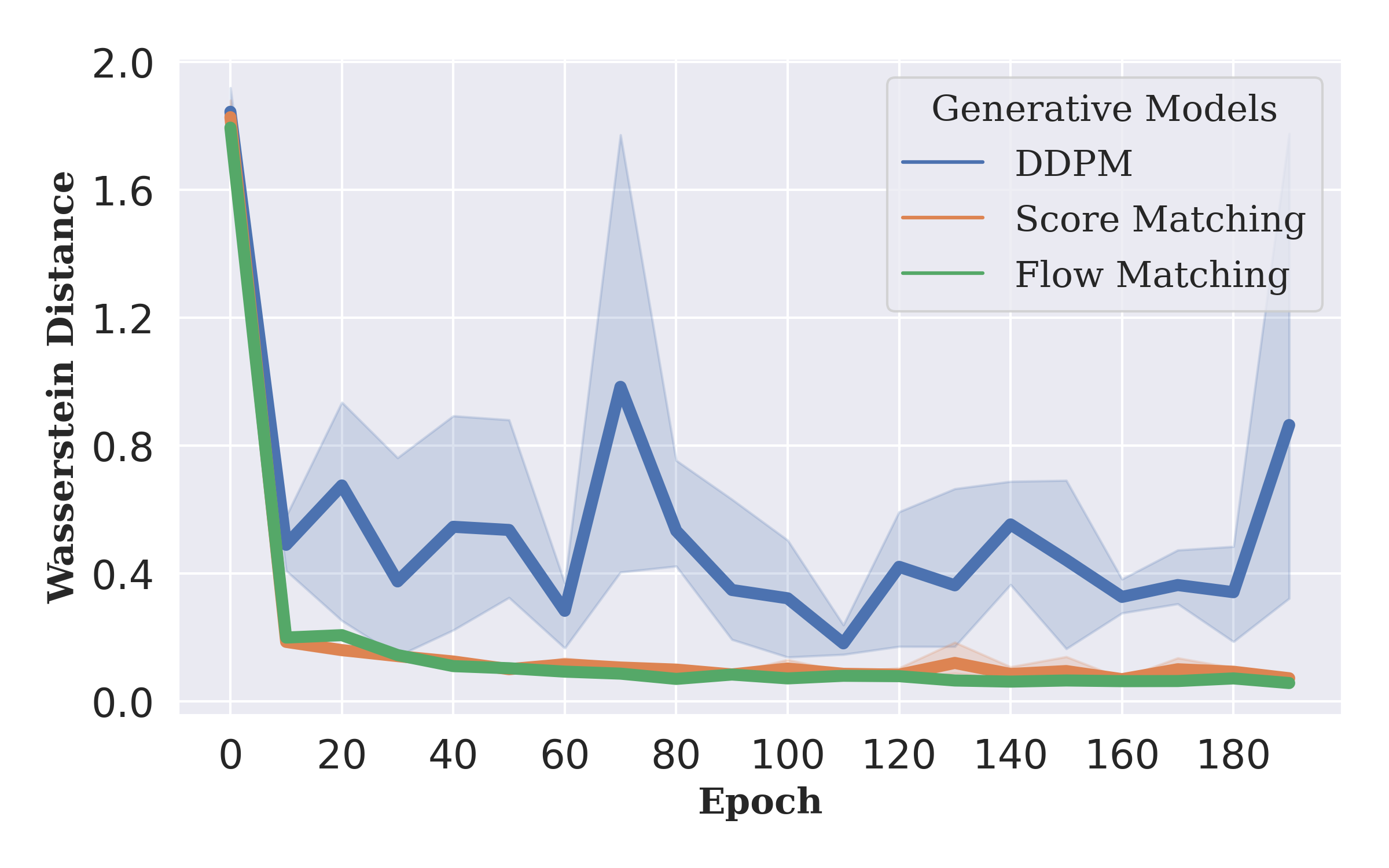}
    }

    \caption{Experiments on SO(3) alignment with different generative modeling methods.}
    \label{rotation}
\end{figure}
\paragraph{Rotation alignment on \texttt{SO(3)} manifold} Different from Euclidean space, \texttt{SO(3)} is a Riemannian manifold \cite{riemannian}. Therefore, the diffusion process (or flow) on SO(3) has been the focus of previous studies, requiring thoughtful design for the following reasons:
(1) Arithmetic operations on Riemannian manifolds are not linearly defined;
(2) Modeling noise  with \texttt{IGSO(3)} rather than a Gaussian distribution;
To abstract the mathematical principles behind rotation alignment from different perspectives, we first provide definitions and formulations in \textcolor{blue}{Appendix. B}. Then we use two sets of synthetic rotation matrices as the target SO(3) distributions, which is visualized using the Euler-angle representation in Figure \ref{rotation} (the range of three Euler angles is set to $[-{\pi}/{2},{\pi}/{2}]$). With increasing training epochs, the \texttt{SO(3)} distributions sampled by the generative models gradually converge to the target distribution, as confirmed by the decreasing 1st-order Wasserstein Distance. 

Compared to DDPM and Score Matching, the curves of Flow Matching on $\mathbb{R}^3$ and \texttt{SO(3)} alignment exhibit better convergence, indicating the superior design performance evaluated in Section \ref{c_benchmark}. 


\section{Datasets}
\paragraph{Unconditional Scaffolding} Following the filtering protocol of FrameDiff\cite{framediff}, we construct a subset of the Protein Data Bank (PDB) \cite{pdb} on August 8, 2021 by selecting monomeric proteins with sequence lengths ranging from 60 to 512 residues and a structural resolution $<5\mathring{A}$. After filtering out any proteins with more than 50\% loops, the dataset is left with 19,703 proteins.

\paragraph{Motif Scaffolding} Motif scaffolding problems consist of sequence and structure constraints on motif(s). We use the same protein subset mentioned above but with lengths range from 60 to 320 (17,576 proteins in total). The motif masks are randomly generated during training, following the mask sampling method described in Genie2 \cite{genie2}. For evaluation, we use a previously published motif scaffolding benchmark Design24 \cite{rfdiffusion} comprising 24 tasks curated from recent publications.

Datasets for training and evaluation are all publicly available at \href{https://dataverse.harvard.edu/dataset.xhtml?persistentId=doi:10.7910/DVN/FADHRC&faces-redirect=true}{\textcolor{blue}{Harvard Dataverse}} (organized as the LMDB cache for efficient and parallel data loading).

\section{Baseline Models}
We ensemble recent open source protein structure design baselines \cite{genie1,genie2,framediff,foldflow,frameflow} in the unified training framework. The renowned method RfDiffusion \cite{rfdiffusion} has received widespread recognition for excelling in \textit{de novo} design, but its training code is unfortunately unavailable, so we report its performance with the official checkpoint as a supplementary reference.

\paragraph{DDPM Methods} combine aspects of the SE(3)-equivariant reasoning machinery of IPA with denoising diffusion probabilistic models to create a diffusion process (conditional or unconditional) over protein structures, including Genie1 \cite{genie1} and Genie2 \cite{genie2}.

\paragraph{Score-Matching Methods} characterize the Brownian motion on the distributions of compact Lie Groups in a amenable form for score matching, and define a forward process on $\texttt{SE(3)}^N$ that allows separation of $\mathbb{R}^3$ and \texttt{SO(3)}, including FrameDiff \cite{framediff} and RfDiffusion \cite{rfdiffusion}.

\paragraph{Flow-Matching Methods} is a family of CNF (continuous normalizing flow) models tailored
for distributions on $\texttt{SE(3)}^N$, which directly regress time-dependent vector fields that
generate probability paths. Representative flow-matching methods are FoldFlow \cite{foldflow} and FrameFlow \cite{frameflow}.

\section{Metrics}
In this section, we briefly introduce the evaluation metrics that will be used to compare different protein structure design methods, including Designability (scTM and scRMSD), Diversity (Pairwise TM), Novelty (Max. TM), Secondary Structure Distribution (Helix percent, Strand percent) and Efficiency (Inference Speed, Model Size etc.). Previous benchmarks~\cite{proteinbench, scaffold-lab} report performance based on different training datasets and experimental setups, making it difficult to attribute performance gains to genuine algorithmic improvement. Protein-\texttt{SE(3)} first retrains each method in a unified framework and then performs evaluation with the same metrics. 

\paragraph{Designability.} A generated protein structure is considered designable if it can be plausibly realized by some protein sequences. To determine whether a model could sample designable proteins, we employ a commonly used pipeline that computes \emph{in silico} self-consistency between generated and predicted structures. First, a generated structure is fed into an inverse folding model (ProteinMPNN~\cite{proteinMPNN}) to produce 8 plausible sequences for the design. Next, structures of proposed sequences are predicted (using ESMFold \cite{esm}) and the consistency of predicted structures with respect to the original generated structure is assessed using structure similarity metrics \cite{tm-score} scTM and scRMSD.

\paragraph{Diversity.} This metric quantifies the diversity of unique structures the design method can generate, ensuring that it produces a broad repertoire of protein backbones rather than merely replicating known folds. We use the average pairwise TM-score of the designable generated samples averaged across lengths as our diversity metric (lower is better).

\paragraph{Novelty.} It evaluates a method’s capacity to explore new structural space beyond known protein folds. We assess the novelty by computing the TM-scores between a designed structure and all structures in the reference PDB dataset, and use the maximum
TM-score obtained to represent novelty (lower is better)). This evaluation is performed using Foldseek \cite{foldseek}, a rapid structural alignment tool. 

\paragraph{Secondary Structure Distribution.} Given protein backbones generated by different methods, the percentages of helix and strand denote the distribution of the secondary structure \cite{ss1,ss2}. A reasonable distribution should approximate the secondary structure distribution of natural proteins, rather than being biased toward alpha helices or beta sheets \cite{genie2,reqflow}.

\paragraph{Efficiency.} Efficiency measures the computational resources (such as the GPU memory and time cost) required to train a \texttt{SE(3)}-based generative model or design a set of proteins. This study reports the training time, inference time, step number, and model size of different methods over our benchmark. Although efficiency may not be a crucial problem compared to diffusion, etc., it is a useful metric to assess the scalability and practicality of the model.

\section{Comprehensive Benchmark}
\label{c_benchmark}
In this section, we present the benchmark developed to systematically evaluate the design performance of the baseline methods. The experiments are organized as follows:
\begin{itemize}[leftmargin=1.5em]
    \item \textbf{Benchmarking unconditional protein structure design across varying lengths.} We benchmark models for unconditional protein structure design with the aforementioned metrics including Designability, Diversity and Novelty. With length-based performance analysis, this benchmark could serve as a standardized assessment for future work.
    \item \textbf{Benchmarking motif scaffolding with Design24.} Motif scaffolding problems consist of structure and sequence constraints on motifs, along with length constraints on target scaffolds. We quantify motif scaffolding capability of baselines (RfDiffusion, Genie2, Frameflow) on the previously published benchmark Design24  \cite{rfdiffusion} with commomly used metrics (scTM and MotifRMSD).
    \item \textbf{In-distribution analysis on secondary structure.} Given the protein backbones generated with different baselines, we record the percentages of the helix and strand, respectively, and visualize the distribution with respect to the percentages. In-distribution analysis assesses how well the model generalizes to natural proteins, rather than biased to a specific structural class.
    \item \textbf{Efficiency Analysis.} To identify potential bottlenecks and foster the development of more efficient and scalable algorithms for protein structure design, we assess the efficiency of integrated methods in terms of training time, inference speed, and model size.
\end{itemize}

 \subsection{Length-based Performance Analysis for Unconditional Scaffolding}
\begin{table}[ht]
    \centering
    \renewcommand{\arraystretch}{1.3}
    \caption{Performance of generative models for unconditional structure design across varying lengths. We highlight the \textbf{best performance} in bold and the \underline{second-best} with the underline. \colorbox{gray!20}{RfDiffusion} is evaluated with the official checkpoint (since the training code is not available). The superscript $*$ indicates that the generation quality does not meet the standard (scTM > 0.5), so Novelty and Diversity are excluded from the comparison.}
    \label{Unconditional_exp}
    \resizebox{\textwidth}{!}{%
    \begin{tabular}{lcccc|cclc}
        \hline
        & \multicolumn{4}{c}{\textbf{Length 100}}& \multicolumn{4}{c}{\textbf{Length 200}}\\
        \cmidrule[0.7pt](rl){2-5} \cmidrule[0.7pt](rl){6-9}
        & \multicolumn{2}{c}{Quality} & \multicolumn{1}{c}{Novelty} & \multicolumn{1}{c}{Diversity} & \multicolumn{2}{c}{Quality} & \multicolumn{1}{c}{Novelty} & \multicolumn{1}{c}{Diversity} \\
        \cmidrule[0.7pt](rl){2-3}  \cmidrule[0.7pt](rl){4-4} \cmidrule[0.7pt](rl){5-5} \cmidrule[0.7pt](rl){6-7}  \cmidrule[0.7pt](rl){8-8} \cmidrule[0.7pt](rl){9-9}
        \textbf{Method}& scTM $\uparrow$ & scRMSD $\downarrow$ & Max TM $\downarrow$ & pairwise TM $\downarrow$ & scTM $\uparrow$ & scRMSD $\downarrow$ & Max TM $\downarrow$ & pairwise TM $\downarrow$ \\
        \hline
        
        Genie1 & 0.89$\pm$0.11 & 1.25$\pm$0.98 & \underline{0.30$\pm$0.09} & \underline{0.35$\pm$0.06} & 0.72$\pm$0.23 & 5.27$\pm$4.60 & 0.23$\pm$0.12 & \underline{0.32$\pm$0.04} \\
        Genie2 & 0.91$\pm$0.08 & 1.04$\pm$0.64 & \textbf{0.29$\pm$0.07} & 0.39$\pm$0.05 & 0.77$\pm$0.19 & 4.01$\pm$3.48 & 0.21$\pm$0.09 & 0.33$\pm$0.03 \\
        FrameDiff & \textbf{0.92$\pm$0.04} & \textbf{0.93$\pm$0.39} & 0.39$\pm$0.13 & 0.37$\pm$0.06 & 0.81$\pm$0.16 & 3.11$\pm$3.08 & 0.34$\pm$0.13 & 0.35$\pm$0.07 \\
        \rowcolor{gray!30} RfDiffusion & 0.97$\pm$0.01 & 0.52$\pm$0.10 & 0.34$\pm$0.13 & 0.39$\pm$0.07 & 0.97$\pm$0.02 & 0.63$\pm$0.15 & 0.31$\pm$0.10 & 0.35$\pm$0.06 \\
        FrameFlow & 0.90$\pm$0.10 & 1.13$\pm$1.03 & 0.38$\pm$0.14 & \textbf{0.33$\pm$0.07} & \textbf{0.94$\pm$0.04} & \textbf{1.24$\pm$0.43} & 0.28$\pm$0.13 & \textbf{0.30$\pm$0.04} \\
        FoldFlow-Base & \underline{0.92$\pm$0.05} & \underline{0.99$\pm$0.36} & 0.36$\pm$0.12 & 0.46$\pm$0.06 & 0.91$\pm$0.04 & \underline{1.50$\pm$0.49} & 0.21$\pm$0.05 & 0.34$\pm$0.04\\
        FoldFlow-OT & 0.91$\pm$0.07 & 1.17$\pm$0.88 & 0.34$\pm$0.08 & 0.43$\pm$0.06 & \underline{0.91$\pm$0.03} & 1.63$\pm$0.60 & \textbf{0.20$\pm$0.04} & 0.34$\pm$0.05 \\
        FoldFlow-SFM & 0.87$\pm$0.06 & 1.39$\pm$0.55 & 0.33$\pm$0.10 & 0.44$\pm$0.06 & 0.82$\pm$0.22 & 3.76$\pm$3.12 & \underline{0.21$\pm$0.05} & 0.35$\pm$0.04 \\
        \hline
                & \multicolumn{4}{c}{\textbf{Length 300}}& \multicolumn{4}{c}{\textbf{Length 500}}\\
        \cmidrule[0.7pt](rl){2-5} \cmidrule[0.7pt](rl){6-9}
        & \multicolumn{2}{c}{Quality} & \multicolumn{1}{c}{Novelty} & \multicolumn{1}{c}{Diversity} & \multicolumn{2}{c}{Quality} & \multicolumn{1}{c}{Novelty} & \multicolumn{1}{c}{Diversity} \\
        \cmidrule[0.7pt](rl){2-3}  \cmidrule[0.7pt](rl){4-4} \cmidrule[0.7pt](rl){5-5} \cmidrule[0.7pt](rl){6-7}  \cmidrule[0.7pt](rl){8-8} \cmidrule[0.7pt](rl){9-9}
        \textbf{Method}& scTM $\uparrow$ & scRMSD $\downarrow$ & Max TM $\downarrow$ & pair TM $\downarrow$ & scTM $\uparrow$ & scRMSD $\downarrow$ & Max TM $\downarrow$ & pair TM $\downarrow$ \\
        \hline
        
        Genie1 & 0.68$\pm$0.19 & 6.57$\pm$3.80 & 0.37$\pm$0.22 & 0.33$\pm$0.07 & 0.47$\pm$0.12$^*$ & 14.39$\pm$4.22 & 0.11$\pm$0.03 & 0.31$\pm$0.04 \\
        Genie2 & 0.64$\pm$0.20 & 7.56$\pm$4.33 & \textbf{0.15$\pm$0.04} & 0.35$\pm$0.03 & 0.46$\pm$0.08$^*$ & 14.28$\pm$2.92 & 0.13$\pm$0.04 & 0.35$\pm$0.04 \\
        FrameDiff & 0.72$\pm$0.13 & 5.40$\pm$2.97 & 0.34$\pm$0.15 & 0.36$\pm$0.09 & \textbf{0.64$\pm$0.19} & \textbf{9.68$\pm$5.17} & 0.20$\pm$0.08 & \underline{0.34$\pm$0.05} \\
        \rowcolor{gray!30} RfDiffusion & 0.94$\pm$0.04 & 1.04$\pm$0.89 & 0.32$\pm$0.12 & 0.38$\pm$0.04 & 0.90$\pm$0.11 & 3.65$\pm$2.95 & 0.24$\pm$0.08 & 0.37$\pm$0.06 \\
        FrameFlow & \textbf{0.90$\pm$0.06} & \textbf{2.00$\pm$0.89} & 0.30$\pm$0.12 & \underline{0.32$\pm$0.09} & 0.56$\pm$0.20 & \underline{11.10$\pm$6.12} & \underline{0.17$\pm$0.08} & 0.35$\pm$0.07 \\
        FoldFlow-Base & \underline{0.87$\pm$0.12} & \underline{2.82$\pm$2.71} & 0.16$\pm$0.04 & 0.33$\pm$0.03 & \underline{0.57$\pm$0.23} & 11.67$\pm$6.70 & \textbf{0.13$\pm$0.03} & \textbf{0.29$\pm$0.03}\\
        FoldFlow-OT & 0.70$\pm$0.21 & 6.01$\pm$4.43 & \underline{0.16$\pm$0.03} & 0.34$\pm$0.04 & 0.38$\pm$0.05$^*$ & 13.35$\pm$3.25 & 0.12$\pm$0.05 & 0.32$\pm$0.04 \\
        FoldFlow-SFM & 0.68$\pm$0.23 & 7.54$\pm$6.53 & 0.17$\pm$0.04 & \textbf{0.30$\pm$0.03} & 0.37$\pm$0.08$^*$ & 15.46$\pm$3.70& 0.12$\pm$0.03 & 0.32$\pm$0.05 \\
        \hline
        
    \end{tabular}}
    
\end{table}

We first retrain baseline models (except RfDiffusion), and then evaluate them for unconditional scaffolding across varying lengths. The results are presented in Table \ref{Unconditional_exp}, which is based on our previously introduced metrics (Quality, Novelty and Diversity). In terms of the Quality metric (scTM and scRMSD), flow-matching based methods (FrameFlow and Foldflow) demonstrates relatively better performance, which is in line with the mathematical analysis in Section \ref{mathematical}. Novelty is also an essential metric to evaluates a method’s capacity to explore new protein structures. With quality constraint (scTM$>$0.5), FoldFlow and Genie2 demonstrate strong performance in generating novel structures. Based on the structural diversity metric, flow-matching models still demonstrates impressive performance across the different chain lengths. It's noteworthy that the performance of all methods across various metrics shows a decline trend as the protein length increases, which suggests that these models generally struggle to create larger proteins due to the increased complexity.

\subsection{Motif Scaffolding Results on Design24}

\begin{figure}[ht]
    \centering

    \subfigure[Average scTM of Motif Scaffolding Designs (Higher is Better) ]{
    \includegraphics[width=0.96\textwidth]{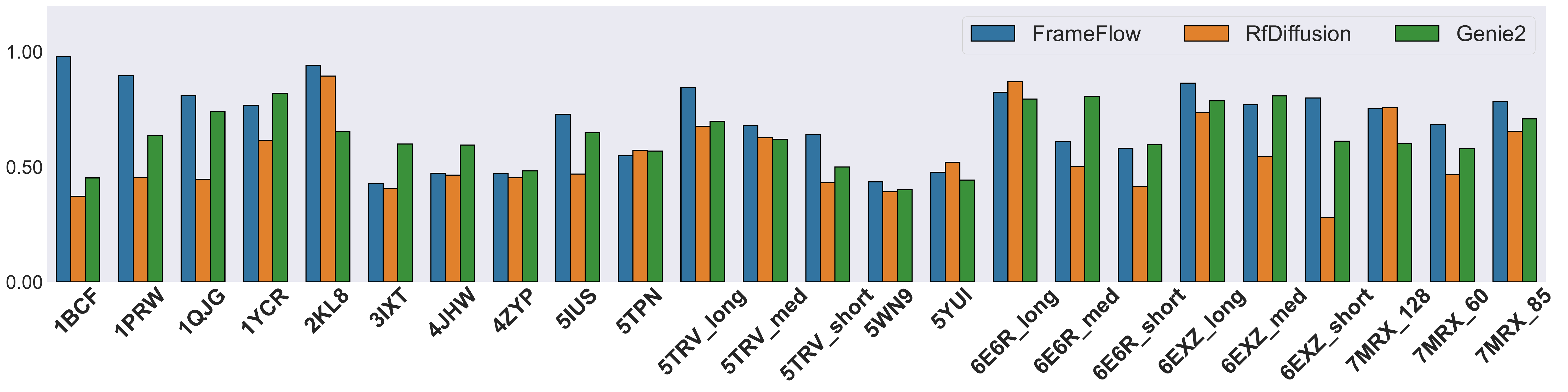}
    }

    \subfigure[Average MotifRMSD of Motif Scaffolding Designs (Lower is Better)]{
    \includegraphics[width=0.96\textwidth]{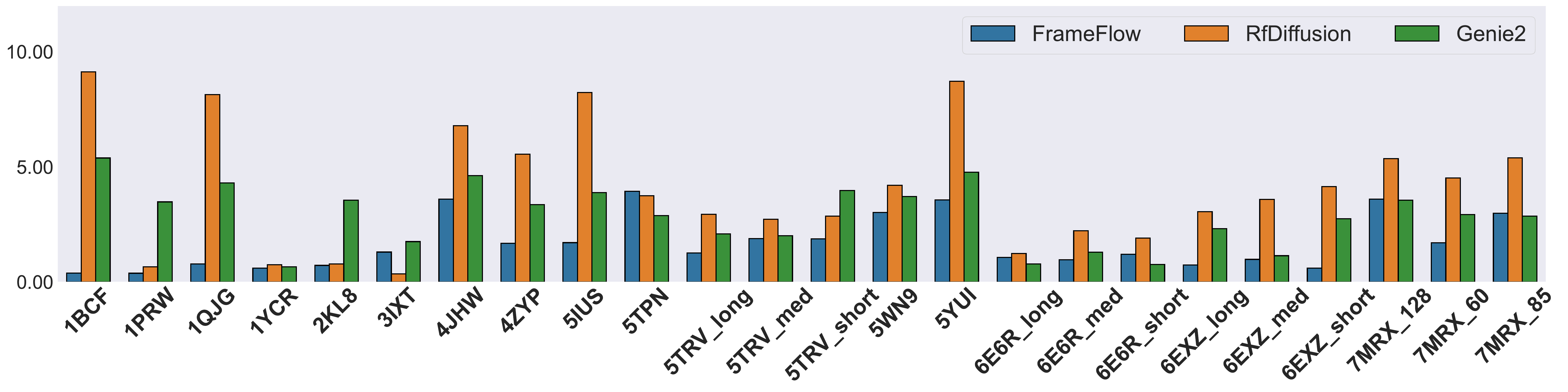}
    }

    \subfigure[Motif Scaffolding Cases (2KL8)]{
    \includegraphics[width=0.15\textwidth]{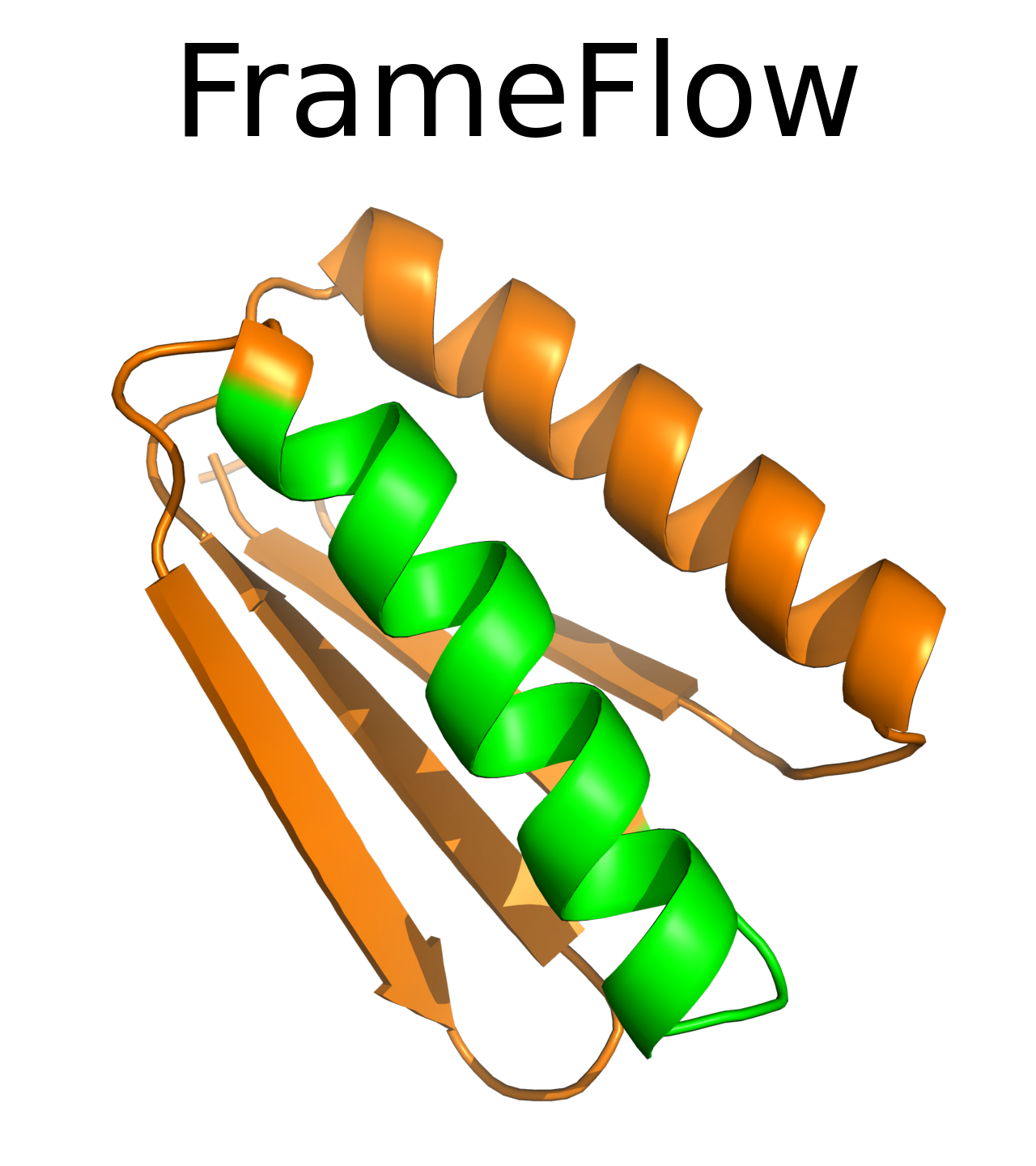}
    \includegraphics[width=0.15\textwidth]{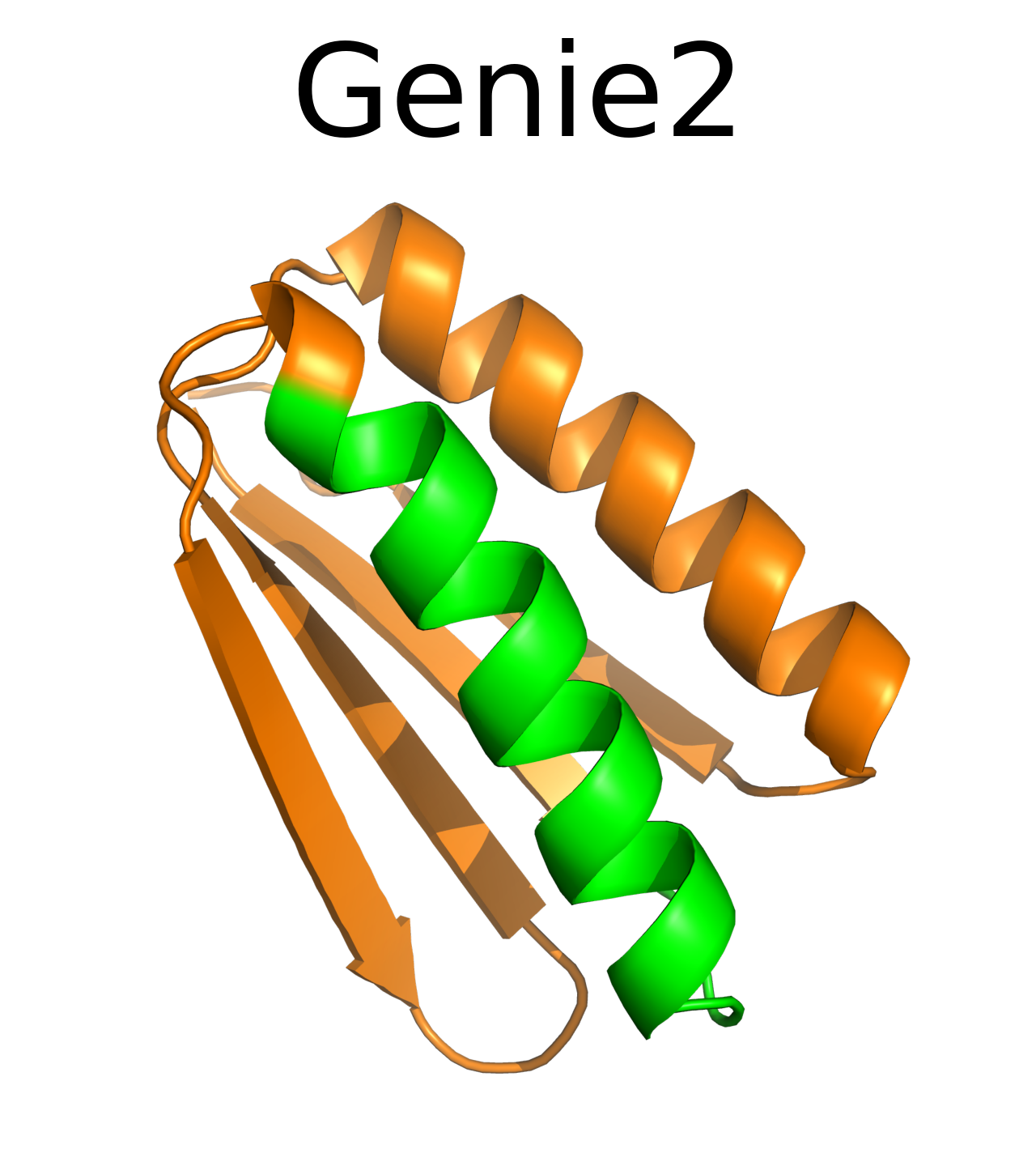}
    \includegraphics[width=0.15\textwidth]{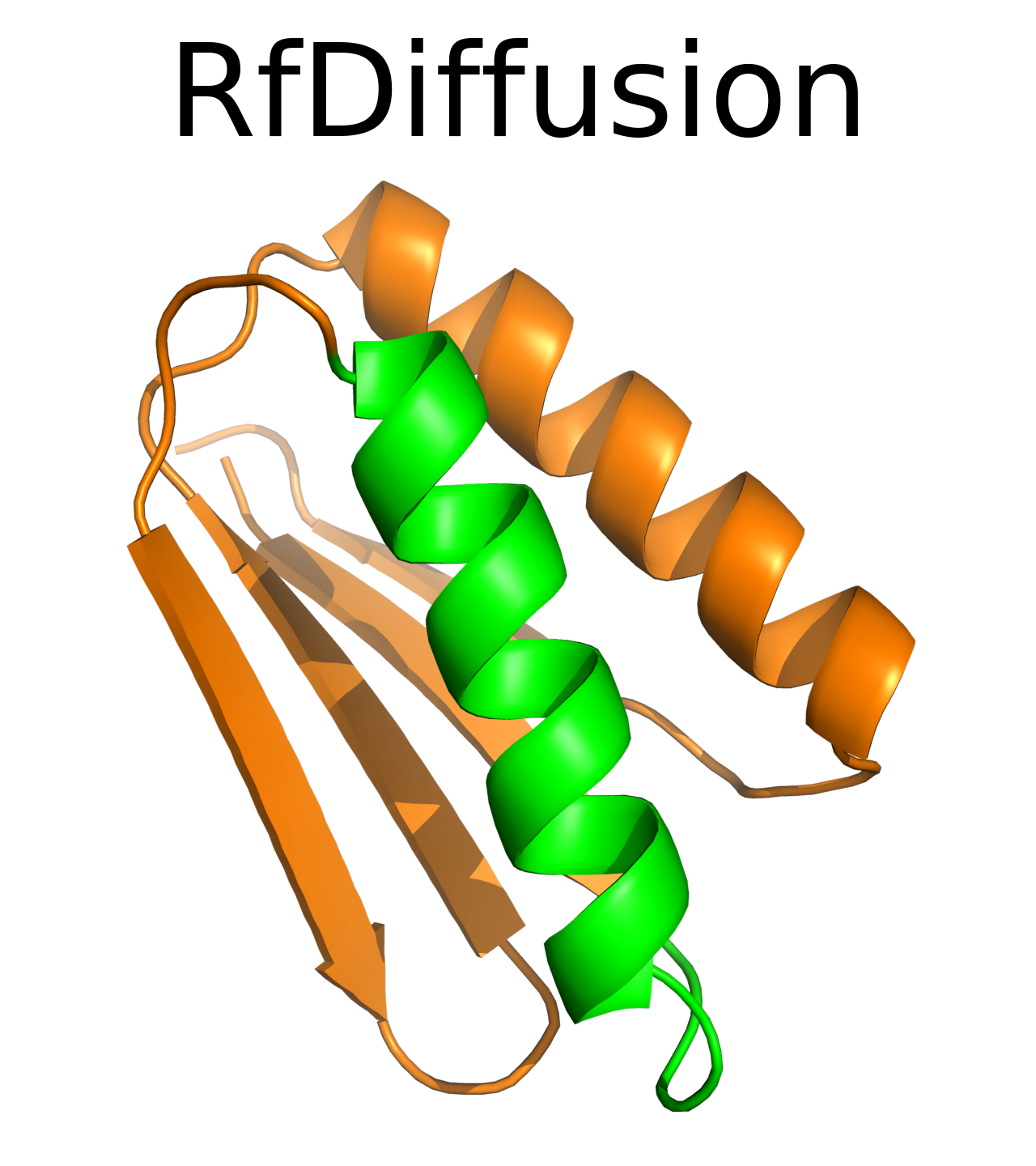}
    }
    \hspace{-1em}
    \subfigure[Motif Scaffolding Cases (5TRV\_med)]{
    \includegraphics[width=0.15\textwidth]{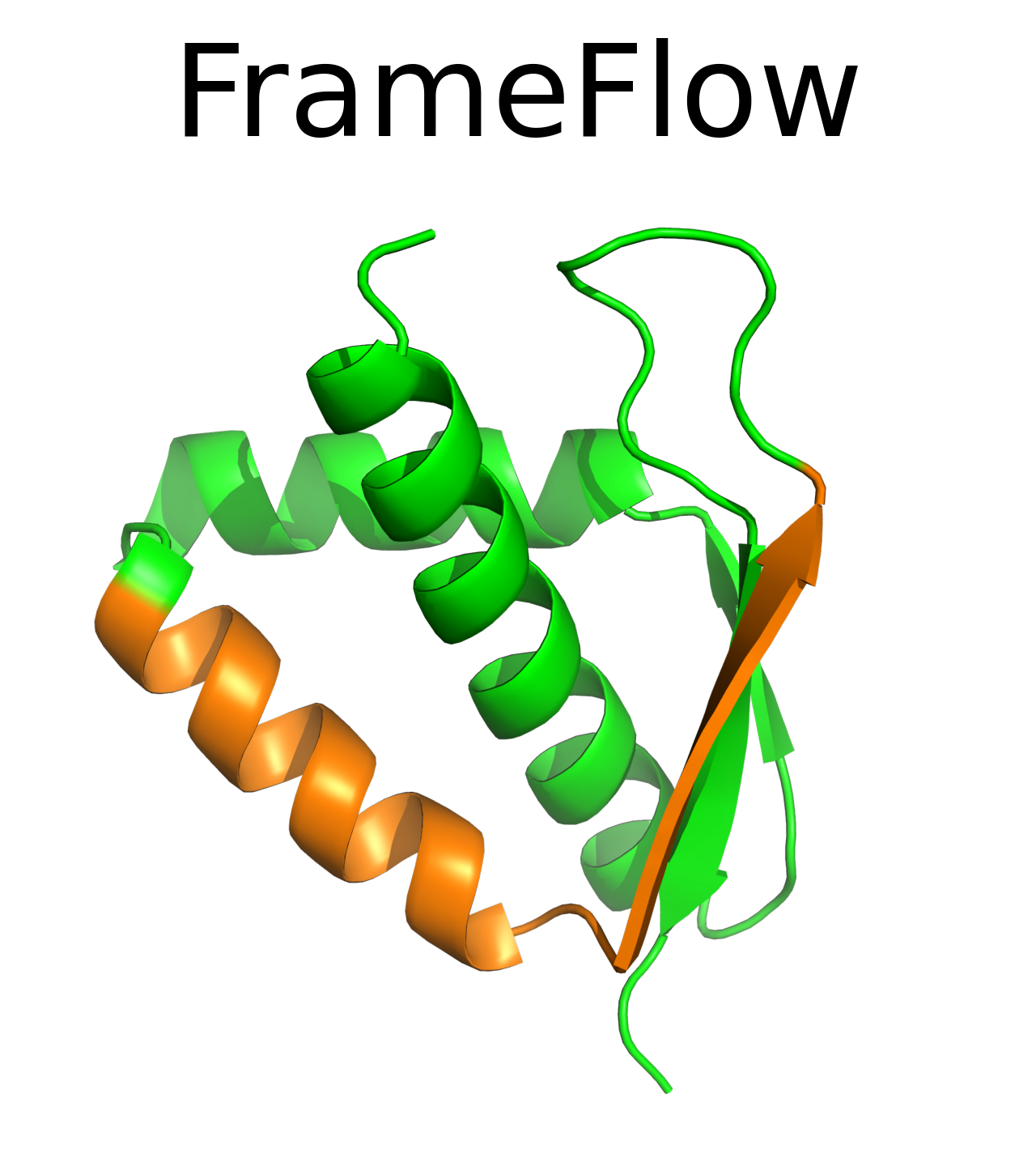}
    \includegraphics[width=0.15\textwidth]{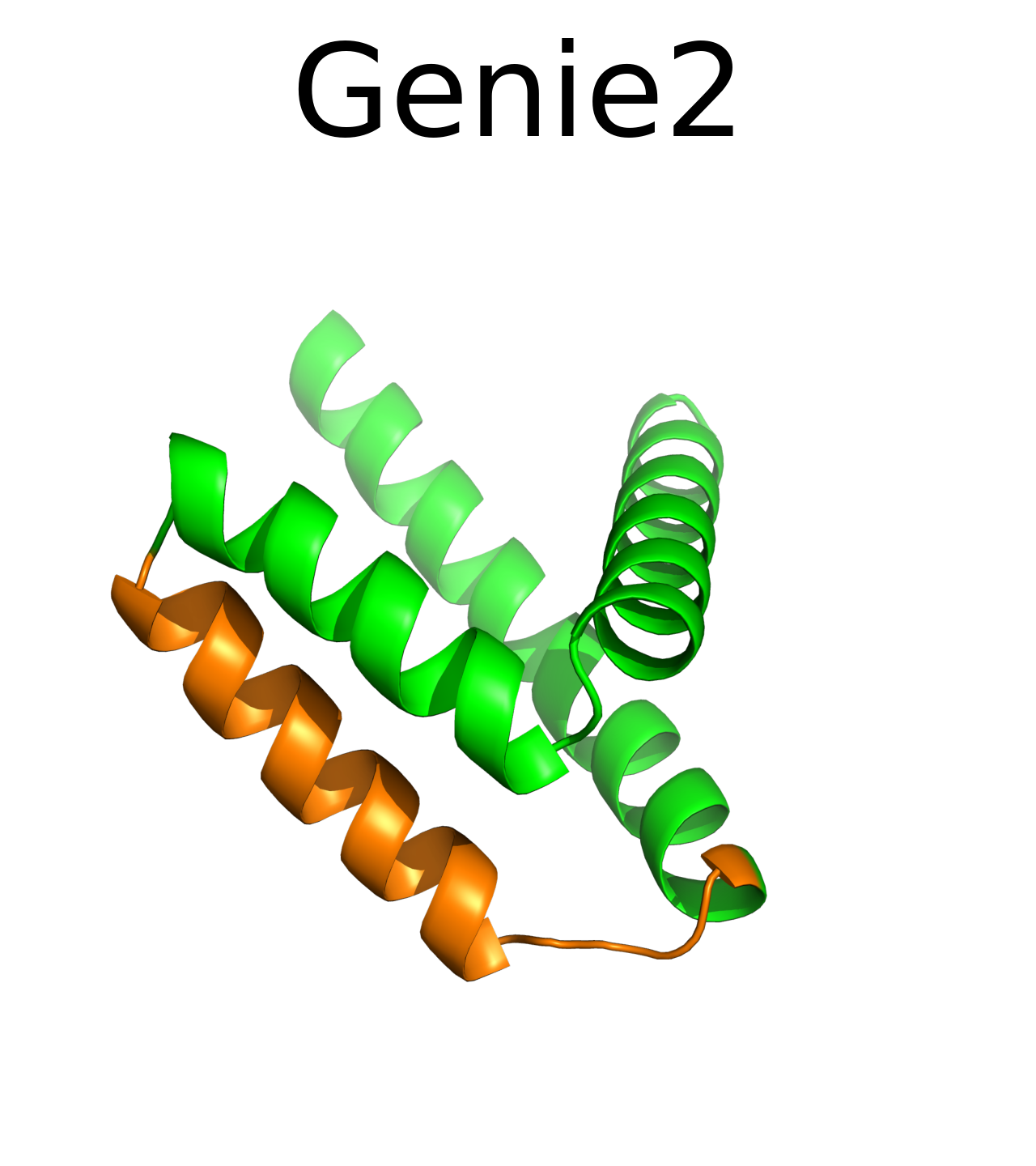}
    \includegraphics[width=0.15\textwidth]{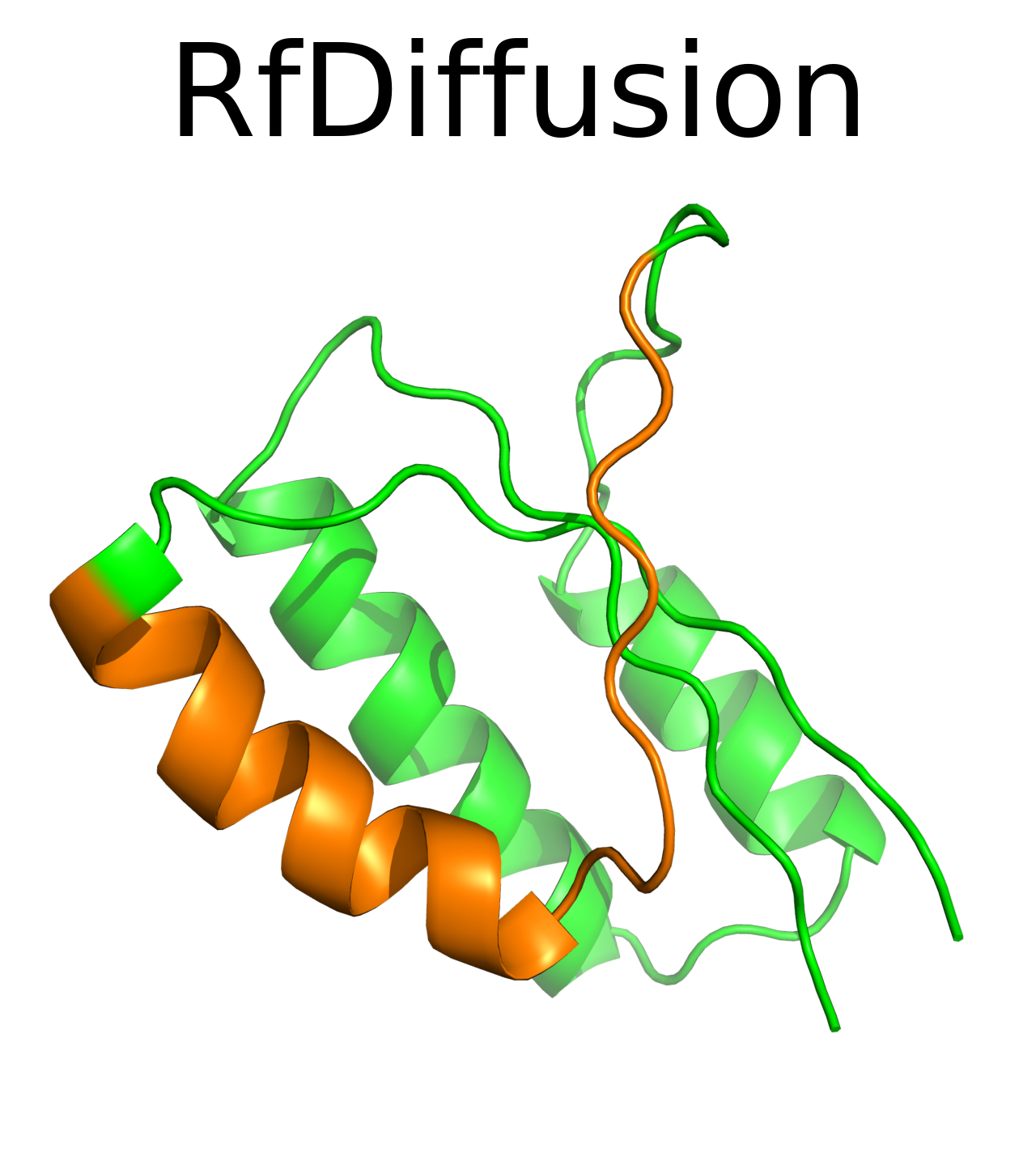}
    }

    \caption{Motif scaffolding results based on Design24. In the case visualization, \textcolor{orange}{motif regions} are colored orange and \textcolor{ForestGreen}{scaffolding regions} green.}
    \label{motif_results}
\end{figure}

We consider three baselines that support the motif scaffolding task. RfDiffusion is evaluated with the official checkpoint (no source code for re-training), while FrameFlow and Genie2 are re-trained with our processed protein dataset with the length ranging from 60 to 320. with 4 scaffolding samples (using the same pipeline for designability evaluation) for each motif, the average scTM and MotifRMSD (scRSMD that only considers the motif region) of three methods are illustrated in Figure \ref{motif_results}.  Frameflow achieves the most designable scaffolds (highest scTM) amongst all methods in 13 out of 24 test motifs compared to Genie2's 7/24 and RfDiffusion's 6/24. For the MotifRMSD metric, FrameFlow still demonstrates superior performance by generating 19 proteins with the lowest MotifRMSD out of 24 tests, compared to the Rfdiffusion's 1/24 and Genie2's 4/24.

\subsection{In-distribuition Analysis on the Secondary Structure}
\label{ss_section}
\begin{figure}[t]
    \centering
    \includegraphics[width=0.9\textwidth]{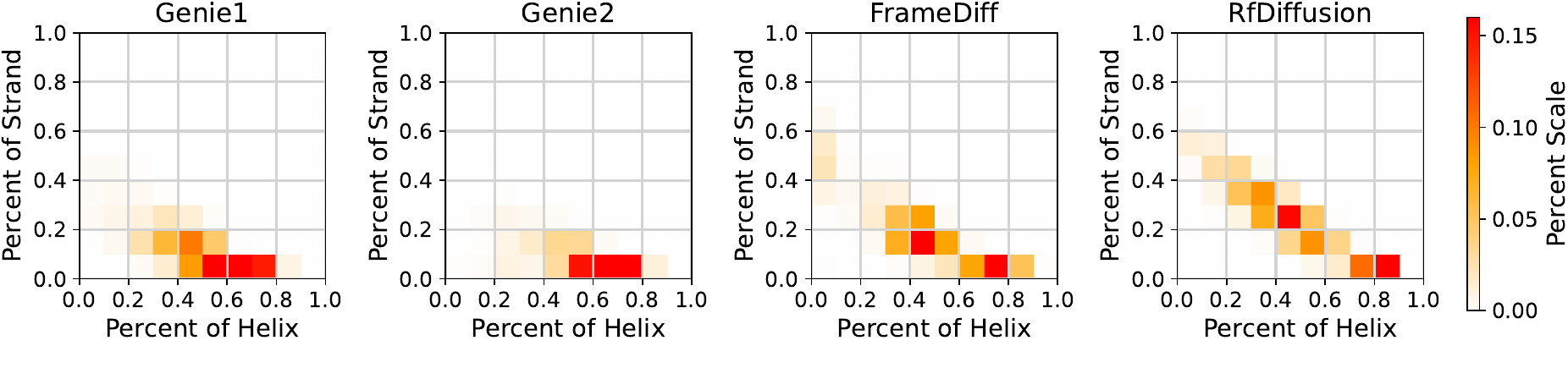}
    \includegraphics[width=0.9\textwidth]{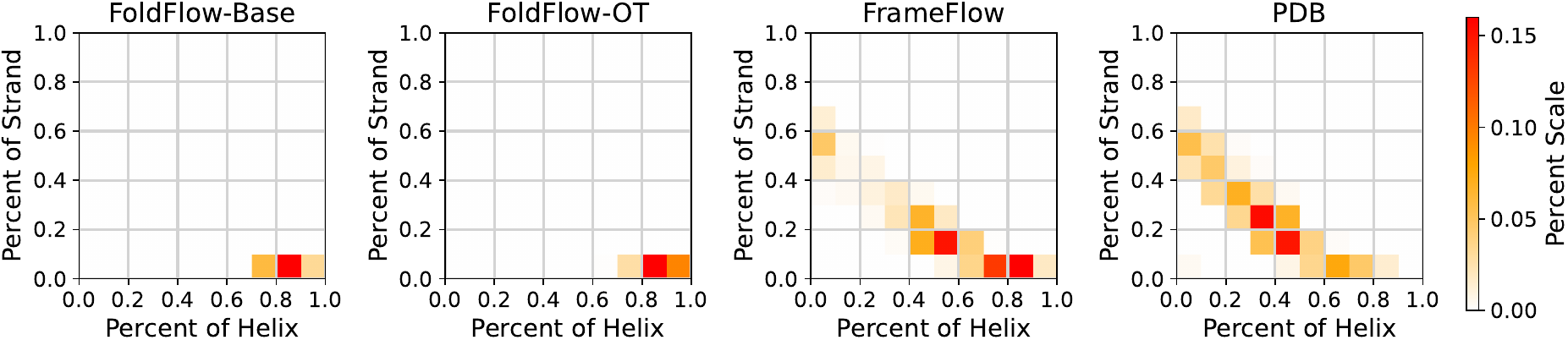}
    \caption{The distribution of secondary structure on unconditionally generated protein structures.}
    \label{ss}
\end{figure}

Given 5 unconditionally generated protein structures with every length ranging from 60 to 320 (1,040 structures in total), we report the secondary structure distribution (the percentages of helix and strand) of each method in Figure \ref{ss}. The proteins generated by official RfDiffusion, re-trained FrameFlow and FrameDiff have more reasonable distributions, which is similar to those randomly sampled from the PDB (bottom right). However, the distributions of FoldFlow, Genie1 and Genie2 differ significantly from the PDB distribution, indicating the risk to generate protein backbones consistently dominated by helical structures. This explains why the performance of these methods is inferior to that of the others in the Quality and Diversity, as shown in the Table \ref{Unconditional_exp}.

\subsection{Computational Efficiency}
\begin{table}[ht]
\renewcommand{\arraystretch}{1.2}
    \centering
    \resizebox{\textwidth}{!}{%
    \begin{tabular}{l|llllll}
    \hline
    Method & Epochs & Training Time  & GPUs & Model Size & Sample Steps & Inference Time \\ \hline
    Genie1~\cite{genie1}  & 100&$\sim$3.0 days  & 2$\times$A100 & 4.10M & 1k & 40 hours \\
    Genie2~\cite{genie2}  & 100&$\sim$3.0 days  & 4$\times$A100 & 15.7M & 1k& 26 hours \\
    FrameDiff~\cite{framediff}& 150&$\sim$4.2 days  & 2$\times$L20 & 18.8M & 100& 31 hours \\ 
    FoldFlow~\cite{foldflow}& 100&$\sim$3.5 days  & 2$\times$L20 & 17.4M & 100& 3.8 hours \\ 
    FrameFlow~\cite{frameflow}& 800&$\sim$2.0 days & 2$\times$L20 & 11.3M & 100 & 1.9 hours  \\
    \hline
    \end{tabular}
    }
    \caption{Efficiency comparison for protein generation models.}
    \label{efficiency}
\end{table}

To encourage developing more efficient and scalable models in the future, we also benchmark the training cost, evaluation cost, model size, and sample steps in Table \ref{efficiency}. Note that the inference time is the total time for generating 1040 proteins with lengths ranging from 60 to 320 (following the setup for in-distribution analysis in Section \ref{ss_section}). Here we found that:
\begin{itemize}[leftmargin=1.5em]
    \item Genie1 and Genie2 consume the most GPU memory during training and the most time for inference, primarily due to the $O(N^3)$ scaling of triangular multiplicative update layers \cite{genie2}.
    \item In terms of inference efficiency, Flow-Matching methods (FrameFlow and FoldFlow) significantly outperform DDPM (Genie1 and Genie2) and Score-Matching (FrameDiff) approaches. This is primarily because flow matching leverages ordinary differential equations (ODEs) to model probability flows \cite{ode, shortcut}, allowing fewer sample steps for efficient generation.
\end{itemize}

\section{Conclusion and Future Work}
We propose Protein-\texttt{SE(3)}, a unified benchmark for \texttt{SE(3)})-based protein generative models, featuring high-level mathematical abstraction to interpret DDPM, Score Matching, and Flow Matching paradigms. It integrates models such as Genie1, Genie2, FrameDiff, RfDiffusion, FoldFlow, and FrameFlow. All methods are evaluated under aligned training settings, enabling fair comparison and rapid prototyping without reliance on explicit protein structures. In the future, we will keep Protein-\texttt{SE(3)}updated in accordance with the latest research, and gradually broaden our scope beyond \texttt{SE(3)}-based structure generation algorithms \cite{chai, all-atom, reqflow}.

\bibliographystyle{plainnat}

\newpage
\appendix
\definecolor{darkblue}{rgb}{0.0, 0.18, 0.65}
\section{Theoretical Formulations of $\mathbb{R}^3$ Alignment}
\label{appendix A}
The translations ($C_\alpha$ positions)  of proteins are defined in the standard $\mathbb{R}^3$ Euclidean space, where the diffusion process (or flow) can be accomplished easily through the previously derived closed form equations. From different perspectives (DDPM~\cite{genie1}, Score Matching~\cite{framediff} and Flow Matching~\cite{foldflow}), we provide details of $\mathbb{R}^3$ alignment implemented in Section \ref{mathematical}.

\subsection{Translation alignment based on DDPM}
\label{r3_align}
Here, we present the formulations of DDPM-based translation alignment following the description of Genie1~\cite{genie1}. Let $\mathbf{x}=[\mathbf{x}_1,\mathbf{x}_2,...,\mathbf{x}_N]$ denote a sequence of $C_\alpha$ coordinates of length $N$. Given a sample $\textbf{x}_0$ from the target distribution over the synthetic data, the forward process iteratively adds Gaussian noise to the sample with a cosine variance schedule $\mathbf{\beta}=[\beta_1,\beta_2,...,\beta_T]$, where the diffusion steps $T$ is set to 1,000:
\begin{equation}
    q(\mathbf{x}_{t} \mid \mathbf{x}_{t-1})=\mathcal{N}(\mathbf{x}_{t} \mid \sqrt{1-\beta_{t}} \mathbf{x}_{t-1}, \beta_{t} \mathbf{I})
\end{equation}
By applying the reparameterization trick to the forward process, we have

\begin{equation}
\begin{split}
        &q(\mathbf{x}_{t} \mid \mathbf{x}_{0})=\mathcal{N}(\mathbf{x}_{t} \mid \sqrt{\bar{\alpha}_t} \mathbf{x}_{0}, (1-\bar{\alpha}_t) \mathbf{I}),\\ &\bar{\alpha}_t=\prod_{i=1}^{t} \alpha_i \ ,\  \alpha_t = 1 - \beta_t
\end{split}
\end{equation}

According to the derivation of DDPM, the reverse process is
modeled with a Gaussian distribution:
\begin{equation}
\begin{split}
    &p(\mathbf{x}_{t-1} \mid \mathbf{x}_{t})=\mathcal{N}(\mathbf{x}_{t-1} \mid \boldsymbol{\mu}_{\theta}(\mathbf{x}_{t}, t), \boldsymbol{\Sigma}_{\theta}(\mathbf{x}_{t}, t) \mathbf{I}),\\
    &\boldsymbol{\mu}_{\theta}(\mathbf{x}_{t}, t)=\frac{1}{\sqrt{\alpha_{t}}}(\mathbf{x}_{t}-\frac{\beta_{t}}{\sqrt{1-\bar{\alpha}_{t}}} \boldsymbol{\epsilon}_{\theta}(\mathbf{x}_{t}, t)) \ , \  \boldsymbol{\Sigma}_{\theta}(\mathbf{x}_{t}, t)=\beta_{t}
\end{split}
\end{equation}
This reverse process requires evaluating  $\boldsymbol{\epsilon}_{\theta}(\mathbf{x}_{t}, t)$ based on MLP layers, which predict the noise added at time step $t$. The loss function is defined as: 
\begin{equation}
\begin{aligned}
L & =\mathbb{E}_{t, \mathbf{x}_{0}, \boldsymbol{\epsilon}}\left[\sum_{i=1}^{N}\left\|\boldsymbol{\epsilon}_{t}-\boldsymbol{\epsilon}_{\theta}(\mathbf{x}_{t}, t)\right\|^{2}\right] \\
& =\mathbb{E}_{t, \mathbf{x}_{0}, \boldsymbol{\epsilon}}\left[\sum_{i=1}^{N}\left\|\boldsymbol{\epsilon}_{t}-\boldsymbol{\epsilon}_{\theta}(\sqrt{\bar{\alpha}_{t}} \mathbf{x}_{0}+\sqrt{1-\bar{\alpha}_{t}} \boldsymbol{\epsilon}_{t}, t)\right\|^{2}\right]
\end{aligned}
\end{equation}
where $\boldsymbol{\epsilon}=\left[\boldsymbol{\epsilon}^{1}, \boldsymbol{\epsilon}^{2}, \cdots, \boldsymbol{\epsilon}^{N}\right]$ and each $\boldsymbol{\epsilon}^{i} \sim \mathcal{N}(\mathbf{0}, \mathbf{I})$.

\subsection{Translation alignment based on Score Matching}
Following FrameDiff~\cite{framediff}, the process of $\mathbb{R}^3$ alignment is modeled as an Ornstein–Uhlenbeck process ( also called VP-SDE~\cite{score-matching}). Still, Let $\mathbf{x}=[\mathbf{x}_1,\mathbf{x}_2,...,\mathbf{x}_N]$ denote a sequence of $C_\alpha$ coordinates of length $N$. Converging geometrically towards Gaussian, the VP-SDE of $\mathbf{x}$ in the $\mathbb{R}^3$ space is:
\begin{equation}
    d \mathbf{x}=-\frac{1}{2} \beta(t) \mathbf{x} d t+\sqrt{\beta(t)} d \mathbf{w}
\end{equation}
where $\beta(t)$ is a non-negative function of $t$ to describe the time-dependent noise schedule, and $\mathbf{w}$ is the standard Wiener process. Its analytical solution is:
\begin{equation}
\begin{split}
        &\mathbf{x}_{t}=\alpha(t) \mathbf{x}_{0}+\sigma(t) \mathbf{z}, \quad \mathbf{z} \sim \mathcal{N}(0, \mathbf{I})\\
        &\alpha(t)=\exp \left(-\frac{1}{2} \int_{0}^{t} \beta(s) d s\right),
        \sigma^{2}(t)=1-\alpha^{2}(t)
\end{split}
\end{equation}
\newpage
To train a score-based model, we want the network $s_\theta(x_t,t)$ to approximate the score function $\nabla_{\mathbf{x}_{t}} \log p\left(\mathbf{x}_{t} \mid \mathbf{x}_{0}\right)$. Note that the conditional distribution is Gaussian:
\begin{equation}
    p\left(\mathbf{x}_{t} \mid \mathbf{x}_{0}\right)=\mathcal{N}\left(\alpha(t) \mathbf{x}_{0}, \sigma^{2}(t) \mathbf{I}\right),
\end{equation}
so the formulation of the true score becomes:
\begin{equation}
    \nabla_{\mathbf{x}_{t}} \log p\left(\mathbf{x}_{t} \mid \mathbf{x}_{0}\right)=-\frac{1}{\sigma^{2}(t)}\left(\mathbf{x}_{t}-\alpha(t) \mathbf{x}_{0}\right).
\end{equation}
With the simple MSE loss function to train the network $s_\theta(x_t,t)$:
\begin{equation}
L = \|s_\theta(x_t,t) - \nabla_{\mathbf{x}_{t}} \log p(\mathbf{x}_{t} \mid \mathbf{x}_{0})\|^2,
\end{equation}
the reverse process from $\mathbf{x}_t$ to $\mathbf{x}_{t-1}$ can be formulated as:
\begin{equation}
    d \mathbf{x}=\left[-\frac{1}{2} \beta(t) \mathbf{x}+\beta(t) s_\theta(x,t)\right] d t+\sqrt{\beta(t)} d \bar{\mathbf{w}},
\end{equation}
where $\bar{\mathbf{w}}$ is denotes Wiener process run backward in time.

\subsection{Translation alignment based on Flow Matching}
Flow Matching in the $\mathbb{R}^3$ space aims to learn a time-dependent velocity field $u_\theta(\mathbf{x},t)$ such that the solution to the following ordinary differential equation (ODE):
\begin{equation}
    \frac{d \mathbf{x}}{d t}=u_{\theta}(\mathbf{x}, t),
\end{equation}
which maps samples from a known base distribution $p_0(\mathbf{x}_0)$ to samples from the target distribution $p_1(\mathbf{x}_1)$ over time $t\in[0,1]$. In the $\mathbb{R}^3$ space, the target velocity at time $t$ is simply:
\begin{equation}
    v_t=\mathbf{x}_1 - \mathbf{x}_0
\end{equation}
The model is trained to predict this velocity via a regression loss:
\begin{equation}
    \mathcal{L}(\theta)=\mathbb{E}_{\mathbf{x}_{0}, \mathbf{x}_{1}, t}\left[\left\|u_{\theta}\left(\mathbf{x}_{t}, t\right)-\left(\mathbf{x}_{1}-\mathbf{x}_{0}\right)\right\|^{2}\right]
\end{equation}
Once the model $u_{\theta}(\mathbf{x}, t)$ is trained, it defines a velocity field over space and time. The sampling process amounts to integrating the learned ODE from a noise sample to a data sample:
\begin{equation}
    \mathbf{x}_{t-1} = \mathbf{x}_{t} +  u_{\theta}\left(\mathbf{x}_{t}, t\right)dt
\end{equation}
Solvers like Euler method and Runge-Kutta can be further used to numerically integrate the ODE for higher speed and accuracy.

\section{Theoretical Formulations of \texttt{SO(3)} Alignment}
Different from Euclidean space, SO(3) is a Riemannian manifold. Therefore, the diffusion process (or probability flow) on SO(3) has been the focus of previous studies, requiring thoughtful design. Here we provide the formulations of rotation alignment in the \texttt{SO(3)} space, which is implemented from different perspectives(DDPM, Score Matching and Flow Matching) in Section \ref{mathematical}. 

\subsection{Preliminaries}
\label{preliminary}
\paragraph{The exponential and logarithmic maps.} Generally speaking, the exponential and logarithmic relate the elements in Lie Group (rotation matrices) to the ones in Lie Algebra (skew-symmetric matrices). The skew-symmetric matrices in Lie Algebra can be specified with a rotation vector $\mathbf{\Phi}$:
\begin{equation}
    \hat{\mathbf{\Phi}} =  \begin{bmatrix}
0 & z & -y \\
-z & 0 & x \\
y & -x & 0
\end{bmatrix}, \ \text{where}\ \mathbf{\Phi}=(x,y,z).
\end{equation}
The magnitude of this vector $\omega=\|\mathbf{\Phi}\|$ represents the angle of rotation, and its direction $\mathbf{n}=\mathbf{\Phi}/\omega$ denotes the axis of rotation. Following the Rodrigues formula, the exponential map (rotation vector $\mathbf{\Phi}$ to rotation matrix $r$) can be simplified to a closed form:
\begin{equation}
    r = \text{exp}(\hat{\mathbf{\Phi}})=cos(\omega)I + sin(\omega)\hat{\mathbf{n}}+(1-cos(\omega)\mathbf{n}\mathbf{n}^T
\end{equation}
Similarly, the matrix logarithm can be expressed using the rotation angle:
\begin{equation}
    \hat{\mathbf{\Phi}}=\text{log}(r)=\frac{\omega}{2sin\omega}(r-r_T),\ \omega=\arccos[(\text{Trace}(r)-1)/2]
\end{equation}

\paragraph{Sampling from \texttt{IGSO3($\mathbf{\mu}$,$\epsilon^2$)}} To form a rotation matrix $r\sim\texttt{IGSO3}(\mu,\epsilon^2)$, we first perform a random sampling from $\texttt{IGSO3}(\mathbf{I},\epsilon^2)$. The axis of rotation $\mathbf{n}$ is uniformly sampled, and the rotation angle $\theta$ is given by the following CDF~\cite{igso3}:
\begin{equation}
    f(\omega)=\sum_{\ell=0}^{\infty}(2 \ell+1) \exp \left(-l(l+1) \epsilon^{2}\right) \frac{\sin ((\ell+1 / 2) \omega)}{\sin (\omega/ 2)},
\end{equation}
which together yield a rotation vector $\mathbf{\Phi}=\omega\mathbf{n}$. Then the rotation vector is shifted by the mean of the distribution to obtain $R=\mathbf{\mu}\text{exp}(\hat{\mathbf{\Phi}})$ as the sampled rotation matrix.

\subsection{Rotation alignment based on DDPM}
Following the previous work \cite{so3_ddpm}, we
implement a \texttt{SO(3)} DDPM model based on several MLP layers for the rotation alignment in Section \ref{mathematical}. With definitions described in Section \ref{preliminary}, the rotation matrix can be scaled by converting them into the Lie algebra (rotation vector), element-wise multiplying by scalar value, and converting back to rotation matrix through exponential map. The matrix scaling operation is defined as:
\begin{equation}
    \lambda(c, r)=\exp (c \log (r)),
\end{equation}
where $\lambda(...)$ is the geodesic flow from $\mathbf{I}$ to $R$ by the amount $c$. Applying these to equations from the original DDPM model we arrive at the following definitions:
\begin{equation}
\begin{aligned}
q\left(r_{t} \mid r_{0}\right) & \left.=\texttt{IGSO3}\left(\lambda\left(\sqrt{\bar{\alpha}_{t}}, r_{0}\right)\right),\left(1-\bar{\alpha}_{t}\right)\right) ;\\
p\left(r_{t-1} \mid r_{t}, r_{0}\right) & =\texttt{IGSO3}\left(\tilde{\mu}\left(r_{t} r_{0}\right), \tilde{\beta}_{t}\right)
\end{aligned}
\end{equation}
and
\begin{equation}
    \tilde{\mu}\left(r_{t}, r_{0}\right)=\lambda\left(\frac{\sqrt{\bar{\alpha}_{t-1}} \beta_{t}}{1-\bar{\alpha}_{t}}, r_{0}\right) \lambda\left(\frac{\sqrt{\alpha_{t-1}}\left(1-\bar{\alpha}_{t-1}\right)}{1-\bar{\alpha}_{t}}, r_{t}\right)
\end{equation}
where $\beta$ and $\alpha$ are schedule values in line with the description in Section \ref{r3_align}. To train the DDPM model $\epsilon_\theta(R_t,t)$ that predicts $R_0$, the loss function is formulated as follows:
\begin{equation}
L = \mathbb{E}\|\epsilon_\theta(r_t,t)r_{0}^T-I\|_F^2,
\end{equation}
where $\|\cdot\|_F$ represents Frobenius norm.

\subsection{Rotation alignment based on Score Matching}
We abstract the mathematical principles behind score matching for \texttt{SO(3)} from the previous work FrameDiff \cite{framediff}. For any $t\in[0,1]$ and $r_0\in\texttt{SO(3)}$, it assumes that $r_t\sim \texttt{IGSO3}(r_0,t)$, which is stated as the Brownian motion on \texttt{SO(3)}. To train a score-based model on \texttt{SO(3)}, we want the network $s_\theta(r_t,t)$ to approximate the score function $\nabla_{r_t} \log p\left(\mathbf{r}_{t} \mid \mathbf{r}_{0}\right)$: 
\begin{equation}
    \nabla_{r_t} \log p\left(r_t \mid r_0\right)=\frac{r_t}{r_0^\top r_t} \log \left\{r_0^\top r_t\right\} \frac{\partial_{\omega} f\left(\omega(r_0^\top r_t), t\right)}{f\left(\omega(r_0^\top r_t), t\right)}
\end{equation}
where $\omega(r)$ is the rotation angle in radians for any $r\in\texttt{SO(3)}$. In our toolkit, we use MSE loss to train the MLP network $s_\theta(r_t,t)$:
\begin{equation}
L = \|s_\theta(r_t,t) - \nabla_{r_t} \log p(r_t \mid r_0)\|^2.
\end{equation}
The reverse process from $r_t$ to $r_{t-1}$ can be formulated as:
\begin{equation}
    r_{t-1}=r_t[g(t)^2s_\theta(r_t,t)dt+g(t)d\bar{\mathbf{w}}],
\end{equation}
where $g(t)$ is the diffusion coefficient and $\bar{\mathbf{w}}$ is the time-reversal Wiener process.

\subsection{Rotation alignment based on Flow Matching}
According to the method described in FoldFlow~\cite{foldflow}, we implemented the SO(3) alignment process based on Flow Matching in our toolkit. Given two rotation matrices $r_0,r_1\in\texttt{SO(3)}$, the geodesic interpolation index by $t$ has the following form:
\begin{equation}
    r_t = r_0\cdot\text{exp}[t\cdot\log(r_0^\top r_1)]
\end{equation}
Flow Matching over \texttt{SO(3)} aims to learn a time-dependent velocity field $u_\theta(\mathbf{r},t)$ such that the solution to the following ordinary differential equation (ODE):
\begin{equation}
    \frac{d \mathbf{r}}{d t}=u_{\theta}(\mathbf{r}, t)
\end{equation}
FoldFlow takes $\log(r_t^\top r_0)$ divided by $t$ as the target velocity at $r_t$. Thus we train the velocity field (MLP layers) $u_\theta(\mathbf{r},t)$ with the following loss function:
\begin{equation}
    \mathcal{L}(\theta)=\mathbb{E}_{\mathbf{r}_{0}, \mathbf{r}_{1}, t} \left\|u_{\theta}\left(\mathbf{r}_{t}, t\right)-\log(r_t^\top r_0)/t\right\|^{2}_{\texttt{SO(3)}}
\end{equation}
where the distance induced by the $\|\cdot\|_\texttt{SO(3)}$ metric is given by:
\begin{equation}
    d_{\texttt{SO(3)}}(r_0,r_1) = \|\log(r_0^\top r1)\|_F,
\end{equation}
With learned velocity field $u_\theta(\mathbf{r},t)$, the reverse process can be written as:
\begin{equation}
    r_{t-1} = r_t\cdot \text{exp}[r_0^\top \cdot u_\theta(r_t,t)dt]
\end{equation}

\section{Broader Impact}
Protein-\texttt{SE(3)} can significantly accelerate algorithmic development in protein modeling by lowering the barrier to entry for researchers from adjacent fields, such as geometry, physics, or machine learning, who may lack domain-specific knowledge in structural biology. Furthermore, the mathematical abstraction toolkit fosters rapid prototyping of new generative models that are not tightly coupled to protein-specific heuristics. In the long term, this could facilitate advances in protein design for therapeutic and industrial applications, such as \textit{de novo} drug discovery, enzyme engineering, and synthetic biology. However, as with all biotechnological tools, there is a potential risk of dual use, particularly in the design of harmful or unintended bio-active molecules. We emphasize that the benchmark itself does not generate novel proteins, but rather provides a platform for studying model behavior under controlled settings. We encourage responsible use of this tool in line with ethical guidelines and open-science best practices.

\end{document}